\documentclass[journal]{IEEEtran}

\usepackage{color}
\usepackage{multirow} 
\usepackage{xspace}
\usepackage{bm} 
\usepackage{bbm}
\usepackage{threeparttable}
\usepackage{CJK}
\usepackage{array} 
\usepackage{balance}
\usepackage{colortbl}
\usepackage{makecell}
\usepackage{graphicx} 
\usepackage{booktabs} 
\usepackage{amsmath}  
\usepackage{amsfonts} 
\usepackage{amssymb} 
\usepackage{utfsym} 
\usepackage{url}

\makeatletter
\DeclareRobustCommand\onedot{\futurelet\@let@token\@onedot}
\def\@onedot{\ifx\@let@token.\else.\null\fi\xspace}
\def\eg{\emph{e.g}\onedot}
\def\ie{\emph{i.e}\onedot}

\def\vs{\emph{vs}\onedot}

\def\etal{\emph{et al}\onedot}
\makeatother

\usepackage{enumitem}
\setlist[itemize]{leftmargin=*}

\begin{document}

\title{Towards Efficient Partially Relevant Video Retrieval with Active Moment Discovering}

\author{Peipei~Song,
        Long~Zhang,
        Long~Lan,
        Weidong~Chen,
        Dan~Guo,~\IEEEmembership{Senior Member,~IEEE},
        Xun~Yang$^*$,
        and~Meng~Wang,~\IEEEmembership{Fellow,~IEEE}
\thanks{This work was supported in part by the National Natural Science Foundation of China (No. 62402471, No. U22A2094, No. 62272435, and No. 62302474), and in part by the China Postdoctoral Science Foundation (No. 2024M763154). This research was also supported by the advanced computing resources provided by the Supercomputing Center of the University of Science and Technology of China (USTC), and the GPU cluster built by MCC Lab of Information Science and Technology Institution, USTC.}
\thanks{Peipei Song, Long Zhang, Weidong Chen, and Xun Yang are with the School of Information Science and Technology, USTC, Hefei 230026, China (e-mail: beta.songpp@gmail.com; dragonzhang@mail.ustc.edu.cn; chenweidong@ustc.edu.cn; xyang21@ustc.edu.cn). Xun Yang is also with the MoE Key Laboratory of Brain-inspired Intelligent Perception and Cognition, USTC. \emph{($^*$Corresponding author: Xun Yang.)}}%
\thanks{Long Lan is with the Institute for Quantum Information, and the State Key Laboratory of High Performance Computing, National University of Defense Technology (NUDT), Changsha, 410073,  China (e-mail: long.lan@nudt.edu.cn).}
\thanks{Dan Guo and Meng Wang are with Key Laboratory of Knowledge Engineering with Big Data (HFUT), Ministry of Education and School of Computer Science and Information Engineering, Hefei University of Technology (HFUT), Hefei, 230601, China, and are with Institute of Artificial Intelligence, Hefei Comprehensive National Science Center, Hefei, 230026, China (e-mail: guodan@hfut.edu.cn; eric.mengwang@gmail.com).}%
}

\maketitle

\begin{abstract}
Partially relevant video retrieval (PRVR) is a practical yet challenging task in text-to-video retrieval, where videos are untrimmed and contain much background content. The pursuit here is of both effective and efficient solutions to capture the partial correspondence between text queries and untrimmed videos. Existing PRVR methods, which typically focus on modeling multi-scale clip representations, however, suffer from content independence and information redundancy, impairing retrieval performance. To overcome these limitations, we propose a simple yet effective approach with active moment discovering (AMDNet). We are committed to discovering video moments that are semantically consistent with their queries. By using learnable span anchors to capture distinct moments and applying masked multi-moment attention to emphasize salient moments while suppressing redundant backgrounds, we achieve more compact and informative video representations. To further enhance moment modeling, we introduce a moment diversity loss to encourage different moments of distinct regions and a moment relevance loss to promote semantically query-relevant moments, which cooperate with a partially relevant retrieval loss for end-to-end optimization. Extensive experiments on two large-scale video datasets (\ie, TVR and ActivityNet Captions) demonstrate the superiority and efficiency of our AMDNet. In particular, AMDNet is about 15.5 times smaller (\#parameters) while 6.0 points higher (SumR) than the up-to-date method GMMFormer on TVR. 
\end{abstract}

\vspace{-0.3cm}
\IEEEpeerreviewmaketitle

\section{Introduction}
With the rapid growth of social media, the text-to-video retrieval (T2VR) task of aligning video candidates with text queries has seen considerable attention and progress \cite{9878037, dong2022reading, dong2021dual, chen2020fine, miech2019howto100m, liu2019use, li2019w2vv++, faghri2017vse++, dong2019dual}. 
However, videos in T2VR datasets are pre-trimmed to be entirely relevant to corresponding text queries, which exists a gap from the real world. In realistic social media or video platforms (e.g., YouTube), a video is usually long-time and contains several moments, among which only one moment is entirely relevant to the corresponding text query \cite{dong2022partially,zhang2023multi,nishimura2023large,chen2023joint}. This congruity causes T2VR models to perform poorly on these untrimmed videos.
To overcome the above-mentioned problem, researchers proposed to solve T2VR in a practical yet challenging scenario, known as partially relevant video retrieval (PRVR)~\cite{dong2022partially,dong2023dual}. 
PRVR aims to retrieve the partially relevant untrimmed videos that contain at least one internal moment related to the given query. 

\begin{figure}[t]
    \centering
    \includegraphics[width=\columnwidth]{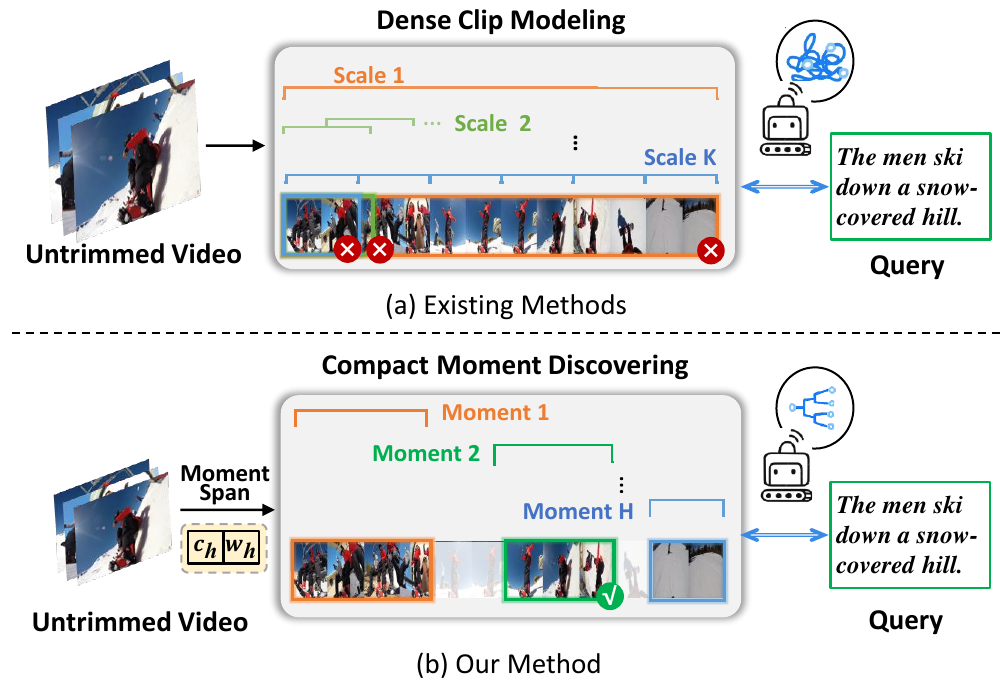}
    \vspace{-0.6cm}
    \caption{Comparison of existing PRVR methods (a) and our method (b). Unlike previous dense clip modeling with content independence and information redundancy, we focus on discovering compact moments in untrimmed videos with learnable moment spans.}
    \label{fig:idea}
\end{figure}

Although remarkable progress has been made on T2VR, the challenging PRVR still remains an unsolved problem due to the partial correspondence between untrimmed video and text query, and the unavailability of moment-query alignment. 
In PRVR, the target video contains plenty of query-irrelevant content. This divergence contradicts the conventional training objective in T2VR models, which aims to establish a mapping between video-text pair \cite{dong2022reading, wang2023gmmformer}. Recall that the video moment retrieval (VMR) task, which aims to retrieve particular moments from a given untrimmed video based on the text query, can be applied to align the text query and moment partition~\cite{10054438, moon2023query, wang2023ms, momentdiff}. However, VMR is limited to a single video rather than large-scale video datasets. {As a result, methods in VMR often benefit from query-dependent video modeling~\cite{moon2023query}, but for PRVR, it becomes extremely time-consuming due to the substantial number of query-video candidates involved.} How to efficiently capture the intrinsic moments in untrimmed videos is one fundamental challenge in PRVR.

Most existing PRVR methods~\cite{dong2022partially,wang2023gmmformer,dong2023dual,jiang2023progressive} focused on modeling dense clip representations to map the partial correspondence between text and video (as shown in Fig. \ref{fig:idea} (a)). They are developed based on the assumption that the relevant moment can be exposed by exhausting clip proposals of different lengths. The dominant approaches typically employ a multi-scale sliding window strategy on consecutive frames to form clip proposals \cite{dong2022partially,wang2023gmmformer}. Then, the text-video similarity is derived from similarities between query embeddings with clip embeddings. However, such dense clip modeling is content-independent and information-redundant. This introduces two inherent bottlenecks: 1) highly overlapping clips have similar semantics, which will confuse the similarity calculation of different query-clip pairs; 2) multi-scale clip construction generates excessive irrelevant clip embeddings and requires a large storage overhead. For instance, the past PRVR method MS-SL \cite{dong2022partially} maintains a total of 528-length clip embeddings, within which only five clips are relevant to corresponding text descriptions on the TVR dataset. 

In this paper, we propose a novel solution leveraging compact moment discovery to deal with the above issues. Our motivation lies in a natural characteristic: a long video contains a few salient moments that are informative and semantically consistent with their queries. Identifying these moments makes the video-query relevance obvious.
As shown in Fig. \ref{fig:idea} (b), we deduce two learnable span anchors (\ie, center and width) from the video, which characterize different moments in an untrimmed video. 
This approach offers two main advantages for solving PRVR. 1) An untrimmed video contains several moments corresponding to different text queries, which the PRVR model should distinguish. By covering different temporal regions with the span anchors, we can extract distinct moments with discriminative semantics, enabling the model to identify the most relevant one for a given text query. 
2) Guided by the learnable moment span, we can construct compact moment-enhanced representations through masked video encoding. For each moment span, the video clips within it are preserved, while those outside it are masked. This strategy emphasizes the portions associated with moments in video features while suppressing irrelevant parts. Consequently, the enhanced video contains less redundant semantics and is more informative for text-to-video retrieval.

\begin{figure*}[t]
    \centering
    \includegraphics[width=\textwidth]{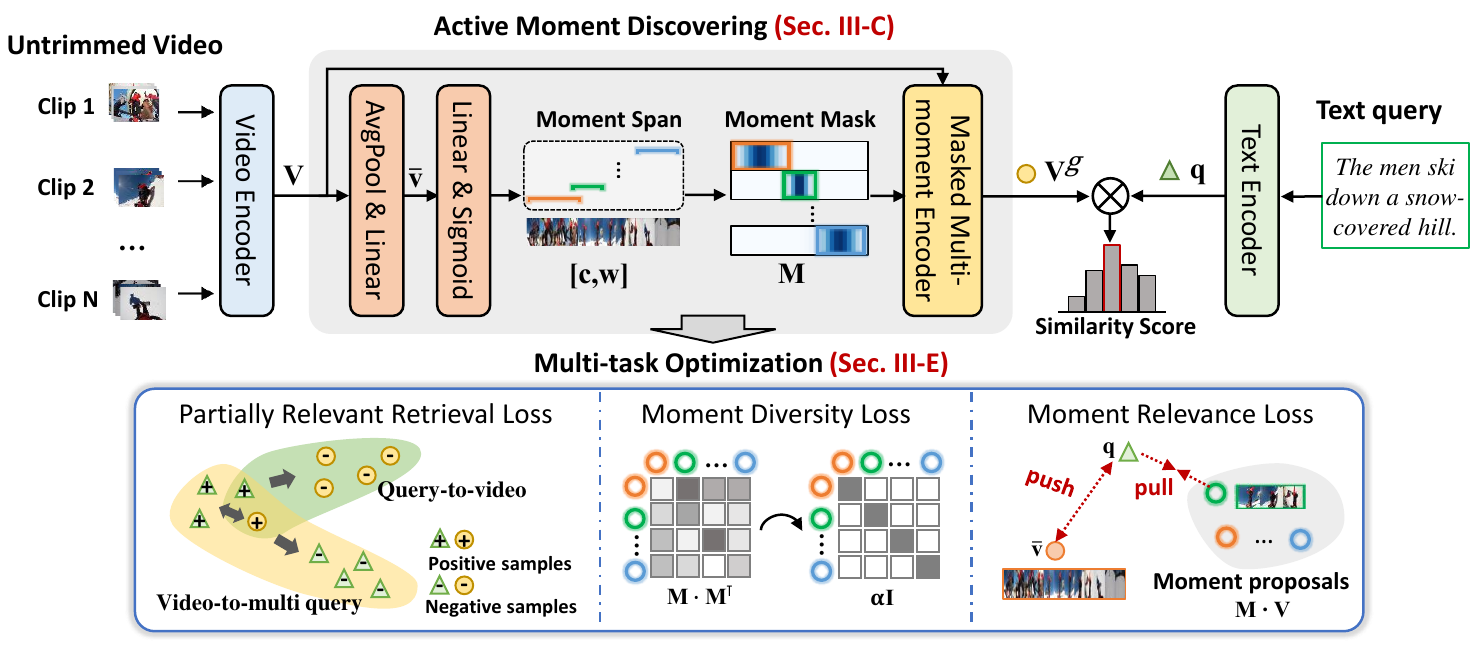}
    \vspace{-0.6cm}
    \caption{An overview of our proposed AMDNet. 
    Given an untrimmed video and query input, we first extract their features ${\bf V}$ and ${\bf q}$. Then, we predict the center and width anchors $[{\bf c},{\bf w}]$ and convert them into a mask matrix ${\bf M}$. ${\bf M}$ is used to modulate the video encodings via masked multi-moment attention and give the moment-enhanced video representations ${\bf V}^g$. Finally, the text-video similarity is obtained by max-pooling the similarity relations between ${\bf V}^g$ and ${\bf q}$. The model is jointly optimized with multi-task losses, including a partially relevant retrieval loss, a moment diversity loss, and a moment relevance loss.}
    \label{fig:main}
\end{figure*}

To implement our idea, we develop a simple yet effective PRVR network with active moment discovering (AMDNet). As shown in Fig. \ref{fig:main}, given an untrimmed video and the corresponding text query, AMDNet first extracts feature embeddings for both the input text query ${\bf q}$ and video frames ${\bf V}$. Subsequently, we predict the center and width anchors conditioned on video, which is then converted into a mask matrix ${\bf M}$ to modulate the video encodings via masked multi-moment attention. In particular, the ${\bf M}$ highlights each informative moment and suppresses the background content outside the moment, we thus obtain the moment-enhanced video representations ${\bf V}^g$.  
{The new ${\bf V}^g$ retains the dimensions of ${\bf V}$ but is enhanced to capture the rich semantics of multiple moment proposals within the untrimmed video.}
Finally, we calculate the text-video similarity by max-pooling the similarity relations between ${\bf V}^g$ and ${\bf q}$.
We optimize the model end-to-end for both cross-modal retrieval and moment discovery tasks, including a partially relevant retrieval loss $\mathcal{L}^{ret}$ to ensure dual retrieval of single videos and multiple queries, a moment diversity loss $\mathcal{L}^{div}$ to encourage less overlap between moments, and a moment relevance loss $\mathcal{L}^{rel}$ to ensure that moments are semantically relevant to their queries.

Overall, our main contributions are as follows:
\begin{itemize}
    \item We propose a new perspective of active moment discovery to address the existing limitations of dense clip modeling in PRVR, in terms of both effectiveness and efficiency. 
    \item We devise a simple yet effective AMDNet, which captures compact and meaningful moments from untrimmed video to improve the partial alignment with queries. A moment relevance loss is designed to ensure semantically sound moment predictions. 
    \item Extensive experiments and ablation studies on two large-scale datasets (\ie, TVR and ActivityNet Captions) demonstrate the superiority and efficiency of our AMDNet. Visualization results further illustrate the effectiveness of moment learning.
\end{itemize}

\section{Related Work}
\subsection{Text-to-video Retrieval}
{
Recent advancements in cross-modal learning, including image-text retrieval \cite{ji2024multi,zhang2023consensus} and referring expression grounding \cite{ji2024progressive}, have sparked growing interest in T2VR tasks by addressing the semantic gap between visual and textual modalities.
} 
Given a textual query, the task of T2VR aims to retrieve relevant videos with the query from a set of pre-trimmed video clips. 
A standard pipeline is to first encode videos and texts to obtain video and sentence representations, and then map them into a common embedding space to measure the cross-modal similarity \cite{dong2018predicting, chen2020fine, li2019w2vv++, faghri2017vse++, li2024momentdiff}. They usually extract video and text features by respective pre-trained unimodal models and learn the cross-modal similarity based on a large amount of video-text pairs. 
With the great success of large-scale image-text pretraining model CLIP \cite{radford2021learning}, most recent works utilize the CLIP encoder for T2VR tasks and achieve state-of-the-art results with an efficient training paradigm \cite{luo2022clip4clip,wu2023cap4video,pei2023clipping,liu2023revisiting,deng2023prompt}. 
However, these T2VR methods above are limited to retrieving pre-trimmed videos, whose semantics are much simpler than videos in current multimedia applications.

\subsection{Partially Relevant Video Retrieval}
The PRVR task \cite{dong2022partially} aims to retrieve untrimmed videos partially relevant to a given query, which is more in line with the real world than T2VR. For PRVR, it is crucial to capture the partial relationship between texts and untrimmed videos. 
Previous studies tackled this task by employing dense matching between the text queries and clip-level video representations. Dong \etal \cite{dong2022partially} proposed multi-scale similarity learning (MS-SL), which constructs the multiple clips from the encoded frame-level representations and computes the cross-modal similarity between the clips and text queries. Afterward, inspired by the capabilities of large-scale multimodal pretraining models, they developed a DL-DKD model \cite{dong2023dual} to distill the text-frame alignment from CLIP. 
{Wang \etal \cite{wang2023gmmformer} utilized multi-scale Gaussian windows to constraint frame interactions of different ranges, and clip features are generated by weighted aggregation of neighboring frames. Then, they proposed GMMFormer v2 \cite{wang2024gmmformer} that introduces a learnable query and weight generator for multi-scale feature aggregation.}
Jiang \etal \cite{jiang2023progressive} deployed dense Gaussian-weighted pooling to summarize the video frames and obtain coarse-grained event representations. 
To improve the efficiency of PRVR, Nishimura \etal \cite{nishimura2023large} proposed splicing a fixed number of adjacent frames as image patches into a super-image. Although resource-friendly, their results show that super-image performs significantly worse than frame sequences.

In this paper, we focus on the PRVR task. Unlike previous practices that traverse all possible clips and yield numerous irrelevant clip embeddings, our proposition involves the usage of learnable span anchors to actively discover prospective moments, which is effective and efficient for the informative grouping of video frames.

\subsection{Video Moment Retrieval}
Unlike PRVR, the VMR task aims to retrieve particular moments from a given single untrimmed video based on the text query~\cite{10054438, moon2023query, wang2023ms, momentdiff,hu2024maskable,yang2022video}. Although the VMR task can be applied to untrimmed videos to align the text and video modalities, it is limited to a single video rather than large-scale video datasets. The video corpus moment retrieval (VCMR) task is an evolution of VMR, which seeks to retrieve moments from a collection of untrimmed videos based on a given query \cite{lei2020tvr,zhang2021video,hou2021conquer,10252035}. VCMR methods usually adopt a two-stage pipeline, where the first stage is to retrieve several candidate videos and the second stage is to retrieve moments from the candidate videos. However, VCMR needs laborious manual annotations of temporal boundaries for every query thus limiting the scalability and practicability in real-world applications.

\subsection{Grouping Video Information Units} 
As consecutive video frames contain highly repetitive information, it is important to encode video into information units to imitate the human behavior of understanding video \cite{song2024emotional,song2023emotion,song2023contextual,guo2024benchmarking}. 
The type of information units varies. There are methods that partition a video into a fixed or adaptive number of segments that consist of successive frames \cite{HRNE,BAE}, select the keyframes that are informative for summarizing the video \cite{PickNet}, gather all the features of video frames at the object-level \cite{OA-BTG,ORG-TRL,SAAT,chen2023weakly,chen2021cascade} or semantic-level \cite{ryu2021semantic}. {Recent works also explore combining audio and visual features \cite{gao2024audio} and performing multi-modal feature interactive fusion \cite{zhang2024descriptive}, further enhancing the video representation.}
For PRVR, how to discover meaningful moment units in videos for text alignment is a to-be-solved issue.

\section{Method}
\subsection{Overview}
PRVR is a challenging task within the field of text-to-video retrieval. Each video in PRVR databases has several moments and is associated with multiple text descriptions, while each text description represents the content of a specific moment in the corresponding video. 
Given a text query $t$, the PRVR task aims to retrieve a video $v$ containing a moment $m^v$ semantically relevant to the given query, from a large corpus of untrimmed videos. It is worth mentioning that the start or end time points of moments are unavailable in PRVR, \ie, the alignment of ($t$, $m^v$) is unavailable. 

A generic PRVR model is to learn a similarity function $S(t, v)$ that scores the similarity between a text query and any video clips \cite{dong2022partially,wang2023gmmformer}. However, 
abundant irrelevant clips seriously affect the accuracy and efficiency of retrieval. With a new perspective, we strive to discover the discriminative moments in the video, thereby potentially learning the similarity of $S(t, m^v)$. 
As shown in Fig. \ref{fig:main}, our method introduces an active moment discovering module. 
It first deduces span anchors from the video and then constructs moment-enhanced video representations ${\bf V}^g$. 
We calculate the similarity of the text-video pair based on the query and moment-enhanced representations. For training, we jointly optimize the model from cross-modal retrieval and moment 
discovery perspectives, with a partially relevant retrieval loss, a moment diversity loss, and a moment relevance loss. The details of each component will be described in the following subsections.

\subsection{Multimodal Representation}
Given an untrimmed video and a natural language query, we first encode them into feature vectors. Following the existing methods \cite{dong2023dual,nishimura2023large,zhang2023multi}, we use CLIP \cite{radford2021learning} as our encoder backbone. 
We first employ a pre-trained CLIP visual encoder to extract frame features of an untrimmed video. Then, to improve the retrieval efficiency, we uniformly sample $N$ feature vectors by mean pooling over the corresponding multiple consecutive frame features and use an FC layer with a ReLU activation to reduce dimension. Finally, we use a transformer block with the learnable positional embedding to capture temporal dependency and get clip features ${\bf V}=\{{\bf v}_n\}_{n=1}^N\in \mathbb{R}^{N\times d}$, where $d$ is the feature dimension. 

For a text query, we employ a pre-trained CLIP text encoder to extract sentence-level features. To connect vision and language domains, we adopt an FC layer with a ReLU activation to embed the text query into the same $d$-dimensional semantic vector space ${\bf q}\in \mathbb{R}^d$ as the video representation ${\bf V}$, which considers semantic context in the sentence.

\subsection{Active Video Moment Discovering}
With the query feature ${\bf q}$ and clip features ${\bf V}$, a native method to obtain the text-video alignment is calculating the feature similarity of ${\bf q}$ and ${\bf V}$ \cite{dong2023dual,dong2022partially}. In this case, each clip ${\bf v}_n$ is treated as a coarse moment candidate for the text query. However, as the empirical finding in \cite{dong2023dual}, primary CLIP features fail to handle the untrimmed videos with mixed query-relevant and query-irrelevant activities. This motivates us to capture informative moments in the untrimmed video that are likely to be described by queries.

\textbf{Moment Span Prediction.} To represent the multiple moments in a video, we employ two span anchors of center ${\bf c}=\{c_h\}_{h=1}^H$ and width ${\bf w}=\{w_h\}_{h=1}^H\in\mathbb{R}^H$, where $0\leqslant {c}_h\leqslant 1$ and $0\leqslant {w}_h\leqslant 1$ indicate the relative positions to the length of the video, $H$ is the pre-defined number of moment proposals within a video. 
Formally, for each video, we predict the moment spans conditioned on the global video semantic $\bar{\bf v}$ as follows:
\begin{gather}
\label{eq:cw}
    \bar{\bf v} = \text{Linear}(\text{AvgPooling}({\bf V}))\in\mathbb{R}^{d},\\
    [{\bf c}, {\bf w}] = \text{sigmoid}(\text{Linear}(\bar{\bf v}))\in\mathbb{R}^{H\times 2}.
\end{gather}
During training, the moment prediction parameters can be learned via backpropagation. 

Then, we prepare a moment mask matrix for subsequent feature calculation. In the experiment, we opt for Gaussian to implement the span-to-mask transformation with reference to \cite{zheng2022weakly}, which is differentiable and can be end-to-end optimized alongside the span generation \cite{jiang2023progressive,wang2023gmmformer,zheng2022weakly}.
Specifically, the moment mask matrix ${\bf M}=\{m_{h,n}|h=1,...,H,n=1,...,N\}\in\mathbb{R}^{H\times N}$ is calculated by:
\begin{gather}
\label{eq:gaussian}
    m_{h,n} = \frac{1}{(\sigma {w}_h)\sqrt{2\pi}}\text{exp}(-\frac{1}{2}\frac{(n/N-{c}_h)^2}{(\sigma {w}_h)^2}),
\end{gather}
where $\sigma$ is a hyperparameter related to the width. 
In $h$-th moment proposal, the mask value $m_{h,n}$ of $n$-clip becomes close to 1 when it is near the center of the moment, and towards 0 as it is further away from the moment.  
Note that the implementation of the span-to-mask transformation is flexible. In Sec. \ref{sec:abla}, we conduct experimental studies to test various transformation strategies, such as Rectangular window and Triangular window \cite{wang2023gmmformer}, our method consistently achieves considerable improvements.

\begin{figure}[t]
    \centering
    \includegraphics[width=\columnwidth]{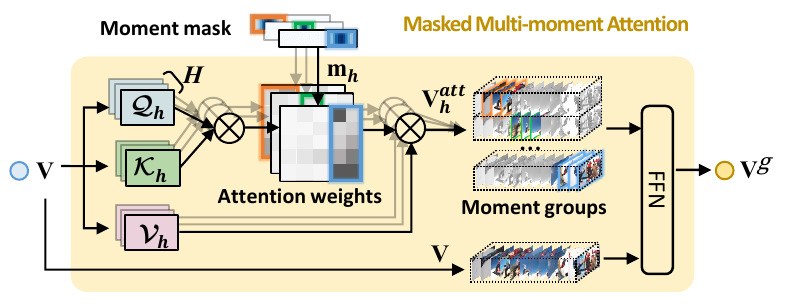}
    \vspace{-0.6cm}
    \caption{Illustration of masked multi-moment attention. It updates the video clip features ${\bf V}$ to moment-enhanced features ${\bf V}^g$ under the guidance of moment mask ${\bf M}$. $H$ is the number of moment proposals in a video.}
    \label{fig:att}
\end{figure}

\textbf{Masked Multi-moment Encoding.}
In order to incorporate the moment clues into the model and obtain moment-enhanced video representations, here we use the moment mask matrix ${\bf M}$ to modulate the video encoding as shown in Fig. \ref{fig:att}.  
Given $H$ moment proposals, we have $H$ sets of queries, keys, and values via three linear transformations, respectively. For $h$-th moment proposal, 
we get query $\mathcal{Q}_h={\bf V}W^q_h$, key $\mathcal{K}_h={\bf V}W^k_h$, and value $\mathcal{V}_h={\bf V}W^v_h$. Then we conduct its mask values ${\bf m}_h=\{m_{h,1},...,m_{h,N}\}$ to perform element-wise product over the query-key attention score, and a softmax function is used to determine attentional distributions over the value. The resulting weight-averaged value forms the summarized video representations ${\bf V}^{att}_h$ for $h$-th moment.
\begin{gather}
    {\bf V}^{att}_h = \text{softmax}({\bf m}_h||_{N} \odot \frac{\mathcal{Q}_h\mathcal{K}_h^{\top}}{\sqrt{d_k}})\mathcal{V}_h\in\mathbb{R}^{N\times d_k},
\end{gather}
where $||_{N}$, $\odot$, and $d_k=d/H$ indicate $N$-time row-wise concatenation, element-wise product, and query/key/value dimensions, respectively.

Finally, we put all the ${\bf V}^{att}_h$ highlighting individual moments and the ${\bf V}$ describing the whole video into the feed-forward network, thereby obtaining the moment-enhanced representations ${\bf V}^g$ of the video. {The ${\bf V}^g$ maintains the full context of ${\bf V}$ while emphasizing moment semantics to promote a comprehensive understanding of the video.} 
\begin{gather}
    {\bf V}^g= \text{FFN}\big([{\bf V}^{att}_1,...,{\bf V}^{att}_H],{\bf V}\big)\in\mathbb{R}^{N\times d},
\end{gather}
where $[,]$ denotes column-wise concatenation. Like the vanilla Transformer block \cite{vaswani2017attention,chen2022multi}, the $\text{FFN}(\cdot)$ combines residual connection, multi-layer perceptron, and layer normalization.

\subsection{Partially Relevant Text-Video Retrieval}
\label{sec:mea}

With the query and video representations, \ie, ${\bf q}$ and ${\bf V}^g$, the similarity between text and video can be measured by feature similarity in the $d$-dimensional embedding space. Considering that a single textual caption can only capture a fragment of the entire video content, we select the maximum similarity between the query feature ${\bf q}$ and any moment-enhanced features ${\bf V}^g$ to represent the similarity of the text-video pair. 
\begin{gather}
S(t, v) = \text{max}\big(\text{sim}({\bf q},{\bf V}^g)\big),   
\label{eq:score}
\end{gather}
where $\text{sim}(\cdot,\cdot)$ is the similarity function in the embedding space \cite{yang2024robust,yang2024learning}, and is implemented by the usual inner product in our experiments.

\subsection{Learning}
Our AMDNet includes three loss items involving cross-modal retrieval and moment discovery tasks: 1) the partially relevant retrieval loss $\mathcal{L}^{ret}$ is used to encourage the dual alignment between most semantically relevant video and text query, 2) the moment diversity loss $\mathcal{L}^{div}$ is used to train the model to produce multiple different moment proposals, and 3) the moment relevance loss $\mathcal{L}^{rel}$ is to ensure the semantic relevance between moment proposals and their queries. Our final loss function is defined as follows to perform joint optimization of all three aforementioned  objectives: 
\begin{gather}
  \mathcal{L} = \lambda_{ret}\mathcal{L}^{ret}+ \lambda_{div}\mathcal{L}^{div}+\lambda_{rel}\mathcal{L}^{rel},
\label{eq:loss}
\end{gather}
where $\lambda_{*}$ are hyperparameters to balance the three losses.

\textbf{Partially Relevant Retrieval Loss.}
For the retrieval part, we adopt an infoNCE loss \cite{miech2020end, zhang2021video} to constrain the dual learning paradigm of text-to-video and video-to-text tasks. Considering the dissimilar granularity between multi-moment videos and single-moment query in PRVR, we compute the loss $\mathcal{L}^{ret}$ for a text-video pair over the mini-batch $\mathcal{B}$ as:
\begin{gather}
\mathcal{L}^{ret} = -\frac{1}{|\mathcal{B}|} \sum_{v \in \mathcal{B}}\bigg \{ \underbrace{\frac{1}{|\mathcal{P}_t|} \sum_{t \in \mathcal{P}_t} {log}(\frac{S(t, v)}{S(t, v)+ \sum\nolimits_{t^{-} \in \mathcal{N}_t }S(t^-, v)})}_\textrm{Video-to-multiquery} \nonumber\\ 
+ \underbrace{{log}(\frac{S(t, v)}{S(t, v) + \sum\nolimits_{v^{-} \in \mathcal{N}_v }S(t, v^-)})}_\textrm{Query-to-video} \bigg \},
\label{eq:nce}
\end{gather}
where $\mathcal{P}_t$ denotes all positive texts of the video $v$ in the mini-batch, $\mathcal{N}_t$ denotes all negative texts of the video $v$ in the mini-batch, while $\mathcal{N}_v$ denotes all negative videos of the query $t$ in the mini-batch. We omit the $exp$ function for brevity. It is worth noting that in the $video$-$to$-$multiquery$ item, we consider all positive texts in $\mathcal{P}_t$ for input video. This encourages similarities between a video and its all positive texts to be increased.

\textbf{Moment Diversity Loss.} 
At the moment discover process, the two span anchors ${\bf c}$ and ${\bf w}$ are learnable and tuned during end-to-end optimization. To encourage the model to capture different moments of distinct regions, we apply a diversity loss $\mathcal{L}^{div}$ as \cite{lin2017structured,zheng2022weakly} to the $H$ moments: 
\begin{gather}
  \mathcal{L}^{div} = ||{\bf M}{\bf M}^{\top}-\alpha {\bf I}||_F^2,
\label{eq:div}
\end{gather}
where ${\bf I}$ is an identity matrix, and $\alpha\in[0,1]$ is a hyperparameter. 
The $\mathcal{L}_{div}$ encourages moments to have less overlap and prevents them from converging to the same center and width. 

\textbf{Moment Relevance Loss.} In addition to diversity, the moments should also be semantically relevant to their queries. However, PRVR datasets lack annotations for the correspondence between queries and moments. 
To this end, we introduce a moment relevance loss $\mathcal{L}^{rel}$ that operates with two sets of relevance scores: one for a high-rank moment and one for the entire video to the query. Specifically, for a query ${\bf q}$, we deem $\text{max}(\text{sim}({\bf q},{\bf V}^m))$ as the positive relevance score for the related moment, where ${\bf V}^m$ represents the RoI features of $H$ moment proposal, defined as ${\bf V}^m={\bf M}\cdot{\bf V}\in\mathbb{R}^{H\times d}$. In order to ensure that the moment group contains only frames highly related to ${\bf q}$, we summarize the entire video as the negative moment candidate. The negative relevance score is calculated using the ${\bf q}$ and the global video feature $\bar{\bf v}$ in Eq. (\ref{eq:cw}). Then, the $\mathcal{L}^{rel}$ is proposed to constrain the relative value of positive and negative relevance scores. The $\mathcal{L}^{rel}$ is formulated as:
\begin{gather}
\label{eq:beta}
    \mathcal{L}^{rel} = \big[\beta+\text{sim}({\bf q},\bar{\bf v})-\text{max}(\text{sim}({\bf q},{\bf V}^m))\big]_+,
\end{gather}
where $\beta$ serves as a margin parameter. $[x]_+=\text{max}(x,0)$. The $\mathcal{L}^{rel}$ decreases with an increase in positive relevance scores relative to the negative relevance scores, thereby encouraging query-related moment prediction.

\section{Experiment}
\subsection{Experimental Setup}
\subsubsection{\textit{\textbf{Dataset}}}
We evaluate our method on two long untrimmed video datasets, \ie, ActivityNet Captions \cite{krishna2017dense} and TVR \cite{lei2020tvr}. Note that moment annotations provided by these datasets are unavailable in the PRVR task.
\textbf{ActivityNet Captions} \cite{krishna2017dense} contains around 20K videos from YouTube, and the average length of videos is around 118 seconds. On average, each video has around 3.7 moments with a corresponding sentence description. For a fair comparison, we adopt the same data partition used in \cite{dong2022partially} with 10,009 and 4,917 videos (\ie, 37,421 and 17,505 annotations) for train and testing, respectively. For ease of reference, we refer to the dataset as ActivityNet.
\textbf{TV show Retrieval (TVR)} \cite{lei2020tvr} contains 21.8K videos collected from 6 TV shows, and the average length of videos is around 76 seconds. Each video is associated with 5 natural language sentences that describe a specific moment in the video. 
Following \cite{dong2022partially}, we utilize 17,435 videos with 87,175 moments for training and 2,179 videos with 10,895 moments for testing.

\begin{table}[t]
    \centering
    \caption{Performance comparison with SOTAs on ActivityNet. 
    DL-DKD-Multi is the extension of DL-DKD with the joint use of CLIP and TCL \cite{yang2022vision}. * indicates our reproduction by official code using CLIP-ViT-B/32 pre-trained weights.}
    \label{tab:act}    
    \renewcommand\arraystretch{1.1}	
    \resizebox{\columnwidth}{!}{ 
    \begin{threeparttable}
    \begin{tabular}{lcccccc} \toprule[2pt]
    \textbf{Method} &\textbf{Venue} &\textbf{R@1} &\textbf{R@5} &\textbf{R@10} &\textbf{R@100} &\textbf{SumR} \\ \hline
    \multicolumn{7}{c}{{T2VR Models}}\\ \hline
    W2VV \cite{dong2018predicting} &TMM'18 & 2.2 & 9.5 & 16.6 & 45.5 & 73.8\\
HTM \cite{miech2019howto100m} &ICCV'19 & 3.7 & 13.7 & 22.3 & 66.2 & 105.9\\
HGR \cite{chen2020fine} &CVPR'20 & 4.0 & 15.0 & 24.8 & 63.2 & 107.0  \\
RIVRL \cite{dong2022reading} &TCSVT'22 & 5.2 & 18.0 & 28.2 & 66.4 & 117.8\\
VSE++ \cite{faghri2017vse++} &BMVC'19 & 4.9 & 17.7 &28.2 & 67.1 & 117.9  \\
DE++ \cite{dong2021dual} &TPAMI'21 & 5.3 & 18.4 & 29.2 & 68.0 & 121.0 \\
DE \cite{dong2019dual} &CVPR'19 & 5.6 & 18.8 & 29.4 & 67.8  & 121.7 \\
W2VV++ \cite{li2019w2vv++} &ACM MM'19 & 5.4 & 18.7 & 29.7 & 68.8 & 122.6 \\
CE \cite{liu2019use} &BMVC'19 & 5.5 & 19.1 & 29.9 & 71.1 & 125.6  \\ 
    CLIP4Clip \cite{luo2022clip4clip} &Neuro.'22 &5.9 &19.3 &30.4 &71.6 &127.3\\ 
    Cap4Video \cite{wu2023cap4video} &CVPR'23 &6.3 &20.4 &30.9 &72.6 &130.2 \\ \hline
    \multicolumn{7}{c}{{VCMR Models w/o Moment Localization}}\\ \hline
    ReLoCLNet \cite{zhang2021video} &SIGIR'21 & 5.7 & 18.9 & 30.0 & 72.0 & 126.6 \\ 
    XML \cite{lei2020tvr} &ECCV'20 & 5.3 & 19.4 & 30.6 & 73.1 & 128.4\\  
    CONQUER \cite{hou2021conquer} &ACM MM'21 &6.5 &20.4 &31.8 &74.3 &133.1 \\ \hline  
    \multicolumn{7}{c}{{PRVR Models}}\\ \hline
    MS-SL \cite{dong2022partially} &ACM MM'22 &7.1 &22.5 &34.7 &75.8 &140.1 \\      
    PEAN \cite{jiang2023progressive} &ICME'23 &7.4 &23.0 &35.5 &75.9 &141.8 \\     
    GMMFormer \cite{wang2023gmmformer}  &AAAI'24 &8.3 &24.9 &36.7 &76.1 &146.0 \\ 
    DL-DKD \cite{dong2023dual} &ICCV'23 &8.0 &25.0 &37.5 &77.1 &147.6 \\     
    DL-DKD-Multi \cite{dong2023dual} &ICCV'23 &8.1 &25.3 &37.7 &77.6 &148.6 \\     
    GMMFormer* \cite{wang2023gmmformer}  &AAAI'24 &10.6 &29.5 &42.6 &79.7 &162.4 \\ 
    MS-SL* \cite{dong2022partially} &ACM MM'22 &11.3 &30.7 &43.5 &81.7 &167.2 \\      
    \rowcolor[RGB]{240,249,254} AMDNet &\textbf{Ours} &\textbf{12.3} &\textbf{32.5} &\textbf{45.9} &\textbf{82.1} &\textbf{172.8} \\ \bottomrule[1pt]
    \end{tabular}
    \end{threeparttable}}
\end{table}

\begin{table}[t]
    \centering
    \caption{Performance comparison with SOTAs on TVR. 
    }
    \label{tab:tvr}
    \renewcommand\arraystretch{1.1}	
   \resizebox{\columnwidth}{!}{ 
    \begin{threeparttable}
    \begin{tabular}{lcccccc} \toprule[1pt]
    \textbf{Method} &\textbf{Venue} &\textbf{R@1} &\textbf{R@5} &\textbf{R@10} &\textbf{R@100} &\textbf{SumR} \\ \hline
    \multicolumn{7}{c}{{T2VR Models}}\\ \hline
    W2VV \cite{dong2018predicting} &TMM'18  & 2.6 & 5.6 & 7.5 & 20.6 & 36.3\\
    HGR \cite{chen2020fine} &CVPR'20  & 1.7 & 4.9 & 8.3 & 35.2 & 50.1  \\
    HTM \cite{miech2019howto100m} &ICCV'19  & 3.8 & 12.0 & 19.1 & 63.2 & 98.2\\
    CE \cite{liu2019use} &BMVC'19 & 3.7 & 12.8 & 20.1 & 64.5 & 101.1  \\
    W2VV++ \cite{li2019w2vv++} &ACM MM'19 & 5.0 & 14.7 & 21.7 & 61.8 & 103.2 \\
    VSE++ \cite{faghri2017vse++} &BMVC'19 & 7.5 & 19.9 & 27.7 & 66.0  & 121.1  \\
    DE \cite{dong2019dual} &CVPR'19  & 7.6 & 20.1 & 28.1 & 67.6  & 123.4 \\
    DE++ \cite{dong2021dual} &TPAMI'21 & 8.8 & 21.9 & 30.2 & 67.4  & 128.3 \\
    RIVRL \cite{dong2022reading} &TCSVT'22 & 9.4 & 23.4 & 32.2 & 70.6 & 135.6\\
    CLIP4Clip \cite{luo2022clip4clip} &Neuro.'22 &9.9 &24.3 &34.3 &72.5 &141.0 \\ 
    Cap4Video \cite{wu2023cap4video} &CVPR'23 &10.3 &26.4 &36.8 &74.0 &147.5 \\ \hline
    \multicolumn{7}{c}{{VCMR Models w/o Moment Localization}}\\ \hline
    XML \cite{lei2020tvr} &ECCV'20 &10.0 & 26.5 & 37.3 & 81.3  & 155.1\\
    ReLoCLNet \cite{zhang2021video} &SIGIR'21 & 10.7 & 28.1 & 38.1 & 80.3  & 157.1 \\  
    CONQUER \cite{hou2021conquer} &ACM MM'21 &11.0 &28.9 &39.6 &81.3 &160.8 \\  \hline  
    \multicolumn{7}{c}{{PRVR Models}}\\ \hline
    MS-SL \cite{dong2022partially} &ACM MM'22 &13.5 &32.1 &43.4 &83.4 &172.4 \\      
    PEAN \cite{jiang2023progressive} &ICME'23 &13.5 &32.8 &44.1 &83.9 &174.2 \\     
    GMMFormer  \cite{wang2023gmmformer}  &AAAI'24 &13.9 &33.3 &44.5 &84.9 &176.6 \\ 
    DL-DKD \cite{dong2023dual} &ICCV'23 &14.4 &34.9 &45.8 &84.9 &179.9 \\ 
    DL-DKD-Multi \cite{dong2023dual} &ICCV'23 &15.1 &35.4 &46.5 &84.5 &181.6 \\     
    MS-SL* \cite{dong2022partially} &ACM MM'22 &17.8 &39.4 &50.7 &88.2 &196.1 \\  
    GMMFormer* \cite{wang2023gmmformer}  &AAAI'24 &18.1 &40.2 &51.7 &\textbf{89.0} &199.1 \\
    \rowcolor[RGB]{240,249,254} AMDNet &\textbf{Ours} &\textbf{19.7} &\textbf{42.4} &\textbf{54.1} &{88.9} &\textbf{205.1} \\ \bottomrule[1pt]
    \end{tabular}
    \end{threeparttable}}
\end{table}

\subsubsection{\textit{\textbf{Evaluation Metric}}}
We comprehensively evaluate the model in terms of retrieval performance and retrieval efficiency. 
\textbf{Performance Metrics.} Following the previous work \cite{dong2022partially}, we utilize the rank-based metrics, namely R@K (K = 1, 5, 10, 100). R@K stands for the fraction of queries that correctly retrieve desired items in the top K of the ranking list. The performance is reported in percentage (\%). The SumR is also utilized as the overall performance, which is defined as the sum of all recall scores. Higher scores indicate better performance.
\textbf{Efficiency Metrics.} We report the total number of parameters for memory consumption and FLOPs for throughput, which computes the total number of floating point operations from visual/textual backbone encodings to video-text similarity calculation. In addition, we measure average runtime and memory usage to complete the retrieval process for a single text query under different database sizes.

\subsubsection{\textit{\textbf{Implementation Details}}}
We uniformly sample $N=32$ clips from each video. For the vision and text encoders, we adopt a Vision Transformer based ViT-B/32 provided by OpenAI\footnote{https://github.com/openai/CLIP}, and encode video frames and query sentences to 512-D features. The dimension of the multimodal feature space is set to $d=256$. The number of moment proposals is set to optimal $H=4$ for ActivityNet and TVR datasets. The hyperparameters in Eq. (\ref{eq:gaussian}) and Eq. (\ref{eq:div}) are empirically set to $\sigma=1/9$ and $\alpha=0.15$ for both datasets. {In Eq. (\ref{eq:beta}), we set $\beta=0.1$ for the ActivityNet dataset and $\beta=0.05$ for the TVR dataset.} The loss coefficients are set to $\lambda_{ret}=0.02$, $\lambda_{div}=1$, and $\lambda_{rel}=1$, which put the three loss terms in the same order of magnitude.
For the model training, we use Adam \cite{kingma2014adam} optimizer with 3$e-$4 learning rate and 128 batch size for 100 epochs. We use the early stop schedule that the model will stop when the evaluated SumR exceeds 10 epochs without promotion as \cite{dong2023dual}.

\begin{figure}[t]
    \centering
    \includegraphics[width=0.9\columnwidth]{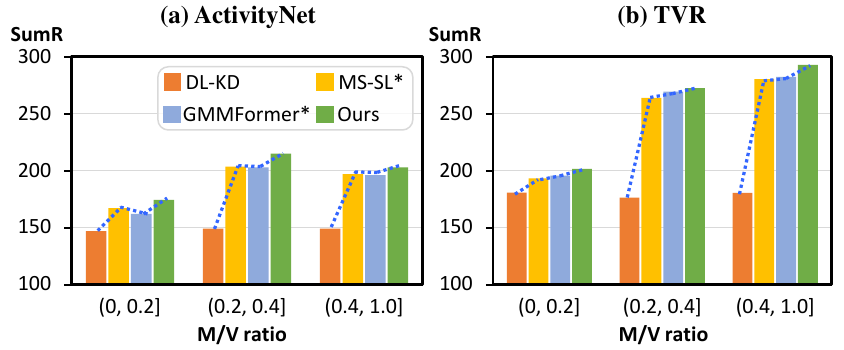}
    \vspace{-0.4cm}
    \caption{Performance on different types of queries. Queries are grouped according to their moment-to-video ratios (M/V). The smaller M/V indicates more challenging queries.}
    \label{fig:mv}
\end{figure}
 
\begin{table}[t]
    \centering
    \caption{Results of the video-to-text retrieval task on ActivityNet and TVR datasets. R@K indicates whether any of the relevant descriptions are ranked in the top K.}
\label{tab:v2t}    
    \renewcommand\arraystretch{1.1}	
\resizebox{\columnwidth}{!}{ 
    \begin{tabular}{lcccccc} \toprule[1pt]
    \textbf{Dataset} &\textbf{Method} &\textbf{R@1} &\textbf{R@5} &\textbf{R@10} &\textbf{R@100} &\textbf{SumR} \\ \hline
    \multirow{3}{*}{ActivityNet} &MS-SL* \cite{dong2022partially} &10.1 &30.7 &46.6 &93.2 &180.5 \\ 
    &GMMFormer* \cite{wang2023gmmformer}  &11.2 &34.9 &51.3 &93.6 &190.9 \\ 
    &AMDNet &\textbf{14.7}  &\textbf{40.8}  &\textbf{56.9}  &\textbf{95.7}  &\textbf{208.1}  \\ \hline
    \multirow{3}{*}{TVR} &GMMFormer* \cite{wang2023gmmformer}  &22.6 &51.4 &66.0 &96.2 &236.3 \\ 
    &MS-SL* \cite{dong2022partially} &27.1 &56.5 &69.2 &96.9 &249.7 \\ 
    &AMDNet &\textbf{26.5} &\textbf{59.6} &\textbf{72.1} &\textbf{97.4} &\textbf{255.6}  \\ 
    \bottomrule[1pt]
    \end{tabular}
    }
\end{table}

\begin{figure}[t]
    \centering
    \includegraphics[width=0.8\columnwidth]{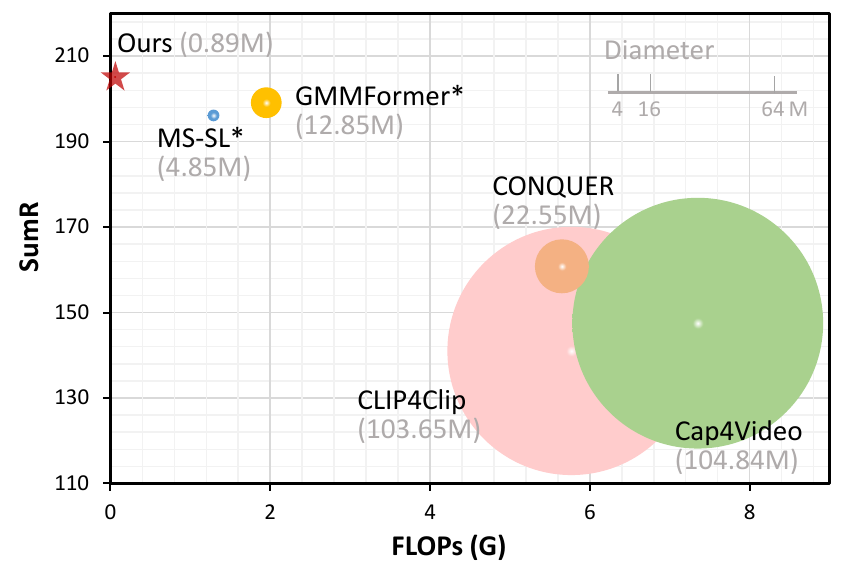}
    \vspace{-0.4cm}
    \caption{{The performance (\ie, SumR), FLOPs, and \# of trainable parameters for various PRVR models on the TVR dataset.} The center of the bubble indicates the value of SumR. The diameter of the bubble or star is proportional to the \#parameters (M) while the horizontal axis indicates the FLOPs (G).}
    \label{fig:size}
\end{figure}

\subsection{Comparison with State-of-the-art Methods}

\subsubsection{\textit{\textbf{Performance Comparison}}}
In Tables \ref{tab:act} and \ref{tab:tvr}, we perform exhaustive comparisons with existing text-to-video retrieval methods on the ActivityNet and TVR datasets, respectively. Related works can be divided into three groups: (1) T2VR models mainly focus on the entire relevance between videos and texts, we compare with various open-source models including the modern CLIP4Clip \cite{luo2022clip4clip} and CapVideo \cite{wu2023cap4video}; 
(2) VCMR models focus on retrieving moments from untrimmed video, 
where a first-stage module is used to retrieve candidate videos followed by a second-stage module to localize specific moments in the candidate videos. The tables report their performance on PRVR datasets by removing moment localization modules; 
(3) PRVR models mainly study clip modeling to learn the partial relevance between videos and texts. Existing works involves multi-scale similarity learning (MS-SL \cite{dong2022partially}), Gaussian-based frame aggregation (PEAN \cite{dong2022partially} and GMMFormer \cite{wang2023gmmformer}), and CLIP-based knowledge distill (DL-DKD \cite{dong2023dual}). In addition, we have re-trained MS-SL and GMMFormer (indicated by *) using the CLIP features.

As shown in Tables \ref{tab:act} and \ref{tab:tvr}, our proposed AMDNet outperforms all the competitor models with clear margins on both datasets. T2VR and VCMR models perform poorly due to their inability to handle partial relevance between videos and texts without moment annotations. 
Compared to PRVR models, we also achieve superior performance. There are the following observations: 
\begin{itemize}
    \item DL-DKD-Multi \cite{dong2023dual} benefits from the multi-teacher distillation based on powerful vision-language pre-training models CLIP and TCL \cite{yang2022vision}. In comparison, our AMDNet using only CLIP weights achieves a considerable SumR improvement of 24.2 and 23.5 on ActivityNet and TVR, respectively. 
    \item When compared with MS-SL* and GMMFormer* which use the same feature extraction backbones with us, our AMDNet improves 8.8\% and 16.0\% on R@1 on ActivityNet relatively. Both MS-SL \cite{dong2022partially} and GMMFormer \cite{wang2023gmmformer} try to discover the consistency between all possible text-clip pairs, where the former builds up clip embeddings by multi-scale sliding windows and the latter adopts multi-scale Gaussian windows. In contrast, the proposed AMDNet performs end-to-end moment modeling and generates a moment-enhanced representation that captures key moments in each video. This representation can be better aligned with the corresponding text query. 
    \item Interestingly, we observe that GMMFormer* gains more improvements from CLIP weights on the TVR than on the ActivityNet compared to its original counterpart. We speculate it is because the ActivityNet contains longer videos than TVR (average 118s \vs 76s per video), which is troublesome for image-based CLIP. However, our proposed model shows strong robustness to distractors and consistently performs the best on both datasets.
\end{itemize}

\begin{table} [tb!]
\caption{Complexity and performance comparisons on TVR and ActivityNet test sets. \textbf{Top:} We measure the average runtime and memory usage of the retrieval process for a single text query under different database sizes on TVR. \textbf{Bottom:} Runtime represents the overall retrieval time on different test sets. * indicates our reproduction by official code using CLIP-ViT-B/32 pre-trained weights. }
\label{tab:efficiency}
\centering 
    \renewcommand\arraystretch{1.1}	
\resizebox{\columnwidth}{!}{ 
\begin{tabular}{lcccccc}
\toprule[1pt]
\textbf{Item} &\textbf{Database Size} & \textbf{500} & \textbf{1,000} & \textbf{1,500} & \textbf{2,000} & \textbf{2,500}\\
\hline
\multirow{3}{*}{Runtime (ms)}  &MS-SL \cite{dong2022partially} & 4.89 & 6.11 & 8.06 & 10.42 & 12.93\\
&GMMFormer \cite{wang2023gmmformer} & {2.68} & {2.93} & {3.40} & {3.94} & {4.56}\\
&AMDNet & \textbf{0.87} & \textbf{1.01} & \textbf{1.09} & \textbf{1.31} & \textbf{1.63}\\
\hline
\multirow{3}{*}{Memory (M)}
&MS-SL \cite{dong2022partially} & 50.02 & 100.04 & 150.06 & 200.08 & 250.11\\
&GMMFormer \cite{wang2023gmmformer} & {2.53} & {5.07} & {7.60} & {10.14} & {12.67}\\
&AMDNet & \textbf{1.62} & \textbf{3.25} & \textbf{4.87} & \textbf{6.50} & \textbf{8.12}\\ \hline\hline
\textbf{Dataset} &\textbf{Method} &\textbf{R@1} &\textbf{R@5} &\textbf{R@10} &\textbf{R@100} &\textbf{Runtime}\\ \hline
\multirow{3}{*}{{TVR}} &{MS-SL*\cite{dong2022partially}  } &{17.8 } &{39.4 } &{50.7 } &{88.2 } &{3,357.66ms} \\  
     &{GMMFormer*\cite{wang2023gmmformer}} &{18.1} &{40.2} &{51.7} &{\textbf{89.0}} &{454.55ms}\\ 
    &{AMDNet} &{\textbf{19.7}} &{\textbf{42.4}} &{\textbf{54.1}} &{{88.9}} &{\textbf{355.85ms}}\\ \hline
    \multirow{3}{*}{{ActivityNet}} &{MS-SL* \cite{dong2022partially}} &{7.1} &{22.5} &{34.7} &{75.8} &{10,610.54ms} \\  
    &{GMMFormer* \cite{wang2023gmmformer}} &{8.3} &{24.9} &{36.7} &{76.1} &{1,335.99ms}\\ 
    &{AMDNet} &{\textbf{12.3}} &{\textbf{32.5}} &{\textbf{45.9}} &{\textbf{82.1}} &{\textbf{521.98ms}}\\
\bottomrule[1pt]
\end{tabular}
}
\end{table}

\subsubsection{\textit{\textbf{Moment-to-video Performance}}}
To gain a more fine-grained comparison, we group the test queries according to their moment-to-video ratio $r$ (M/V) \cite{dong2022partially}, defined as their relevant moment’s length ratio in the entire video. The smaller M/V indicates less relevant content while more irrelevant content in the target video to the query, showing more challenging of the corresponding queries.
As with \cite{dong2023dual}, we compute the sumR scores for three M/V settings, where the moments are short ($r \in$ (0, 0, 2]), middle ($r \in$ (0.2, 0.4]), and long ($r \in$ (0.4, 1.0]). 
Fig. \ref{fig:mv} presents the M/V results on ActivieyNet and TVR. Our proposed model consistently performs the best, which again verifies its effectiveness.

\subsubsection{\textit{\textbf{Evaluation on Video-to-text Retrieval}}}
In addition, we report the performance of GMMFormer \cite{wang2023gmmformer}, MS-SL \cite{dong2022partially}, and our AMDNet on both datasets on the video-to-text task.
As shown in Table \ref{tab:v2t}, our model also demonstrates significant improvements to comparison models across all metrics on both datasets, \eg, on ActivityNet, we improve the SumR from 180.5 and 190.9 to 208.1. This suggests that our compact video moment learning facilitates dual correspondence between long videos and multiple texts.

\subsection{Efficiency Comparison}
In Fig. \ref{fig:size}, we compare some competitive models in terms of FLOPs and model parameters. Following the convention in previous works \cite{dong2019dual,wang2023gmmformer}, we report only the number of trainable parameters and floating point operations from visual/textual backbone encodings to video-text similarity calculation.
The proposed AMDNet is a lightweight model with merely 0.89M parameters. It achieves the best performance (6.0 SumR better than GMMFormer* \cite{wang2023gmmformer}) while the smallest FLOPs (32.25 times smaller than MS-SL* \cite{dong2022partially}). This demonstrates that our considerable performance advantage is independent of explosive parameter increase.

We further measure the runtime and memory usage of the compared methods during inference on the test set. To make the experiment setting close to real-world scenarios and for fair comparisons, we only monitor space and time consumption for the ranking procedure. Compared to MS-SL \cite{dong2022partially} and GMMFormer \cite{wang2023gmmformer}, our proposed method does not require dense modeling of video clips or the score fusion of frame-branch and clip-branch. As shown in Table \ref{tab:efficiency} (Top), our model is about 5.6/3.1 times faster than MS-SL/GMMFormer and has a storage overhead 30.9/1.6 times smaller than MS-SL/GMMFormer on 500 videos. 
As the video database size increases from 500 to 2,500, the retrieval time only increases from 0.87ms to 1.63ms. Our model shows high efficiency for applications. Meanwhile, AMDNet demonstrates a clear advantage in the trade-off between retrieval time and accuracy, as shown in Table \ref{tab:efficiency} (Bottom).  
As we scale from TVR (2,179 videos) to ActivityNet (4,917 test videos), AMDNet effectively maintains its balance of speed and accuracy even as dataset size increases.

\begin{table}[t]
    \centering
    \caption{Ablation studies on the ActivityNet dataset. Removing  ${\bf V}^g$ stands for removing the active moment discovering module, where ${\bf V}^g$ degenerates to base ${\bf V}$.}
\label{tab:wo}
    \renewcommand\arraystretch{1.1}	
    \resizebox{\columnwidth}{!}{ 
    \begin{tabular}{ccc|ccccc} \toprule[1pt]
    \textbf{${\bf V}^g$} &\textbf{$\mathcal{L}^{div}$} &\textbf{$\mathcal{L}^{rel}$} &\textbf{R@1} &\textbf{R@5} &\textbf{R@10} &\textbf{R@100} &\textbf{SumR} \\ \hline
    $\usym{2717}$ &$\usym{2717}$ &$\usym{2717}$ &10.4 &30.5 &43.4 &80.8 &165.1 \\
    $\usym{2713}$ &$\usym{2717}$ &$\usym{2717}$ &11.4 &31.5 &44.3 &81.5 &168.7 \\
    $\usym{2713}$ &$\usym{2713}$ &$\usym{2717}$ &11.6 &31.9 &44.6 &81.7 &169.9 \\
    $\usym{2713}$ &$\usym{2713}$ &$\usym{2713}$     &\textbf{12.3} &\textbf{32.5} &\textbf{45.9} &\textbf{82.1} &\textbf{172.8} \\ 
    \bottomrule[1pt]
    \end{tabular}
    \footnotesize
    }
\vspace{-0.3cm}
\end{table}

\begin{table}[t]
    \centering
    \caption{The effects of the number of moment proposals $H$. Larger H helps to discover all possible moments, but also causes short and incomplete moments. The optimal values are $H$ = 4 on ActivityNet and TVR.}
\label{tab:numK}    
    \renewcommand\arraystretch{1.1}	
\resizebox{\columnwidth}{!}{ 
    \begin{tabular}{lcccccc} \toprule[1pt]
    \textbf{Dataset} &\textbf{Method} &\textbf{R@1} &\textbf{R@5} &\textbf{R@10} &\textbf{R@100} &\textbf{SumR} \\ \hline
    \multirow{4}{*}{ActivityNet} &$H$=1 &11.3 &31.7 &44.5 &81.7 &169.1 \\
    &$H$=2 &11.1 &32.4 &45.4 &81.9 &170.9 \\
    &$H$=4 &\textbf{12.3} &\textbf{32.5} &\textbf{45.9} &\textbf{82.1} &\textbf{172.8} \\
    &$H$=8 &11.6 &32.0 &44.7 &81.8 &170.1 \\ \hline
    \multirow{4}{*}{TVR} &$H$=1 &18.9  &41.6  &52.7  &88.4  &201.6 \\
    &$H$=2 &19.3  &41.9  &52.9  &88.8  &202.9 \\
    &$H$=4 &\textbf{19.7} &\textbf{42.4} &\textbf{54.1} &\textbf{88.9} &\textbf{205.1} \\
    &$H$=8 &19.0  &41.5  &53.0  &88.5  &202.0 \\
    \bottomrule[1pt]
    \end{tabular}
    }
\end{table}

\subsection{Ablation Study}
\label{sec:abla}

\subsubsection{\textit{\textbf{Main Components}}}
In Table \ref{tab:wo}, we conduct ablation studies on the full AMDNet, w.r.t. the moment-enhanced representations ${\bf V}^g$, the moment diversity loss $\mathcal{L}^{div}$, and the moment relevance loss $\mathcal{L}^{rel}$. It can be found that starting with a pure baseline (Line 1), AMDNet gains 3.6 on SumR
by replacing the clip-level representations ${\bf V}$ with the ${\bf V}^g$ (Line 2). 
Adding moment diversity loss further brings an improvement to 4.8 (Line 3) compared to the baseline. 
By jointly using our designed moment encoding, moment diversity loss, and moment relevance loss, AMDNet acquires an improvement of 7.7 on SumR (Line 4). 
These ablations demonstrate the effectiveness of our designed components in improving retrieval performance.

\begin{figure}[t]
    \centering
    \includegraphics[width=0.9\columnwidth]{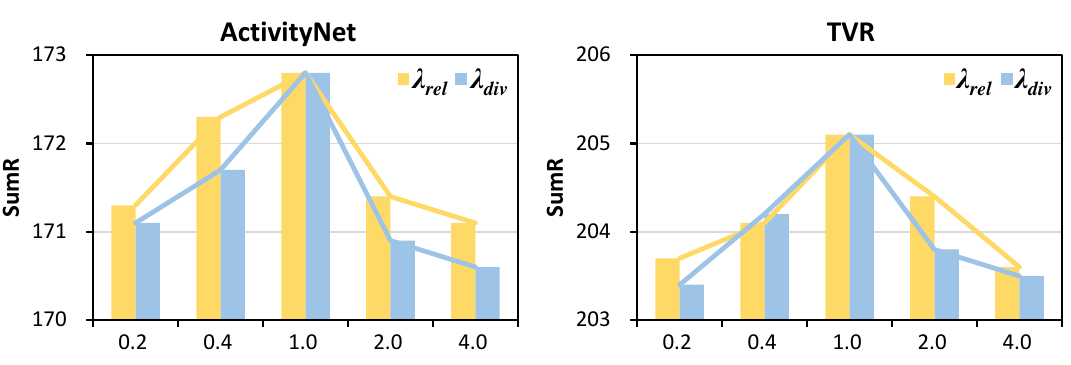}
    \vspace{-0.4cm}
    \caption{Effects of the hyperparameters $\lambda_{div}$ and $\lambda_{rel}$ in terms of SumR metric on ActivityNet and TVR datasets. $\lambda_{ret}$ is fixed to 0.02 for fair comparisons. The performance peaks at $\lambda_{div}=1$ and $\lambda_{rel}=1$.
    }
    \label{fig:para}
\end{figure}

\begin{figure*}[t]
    \centering
    \includegraphics[width=0.9\textwidth]{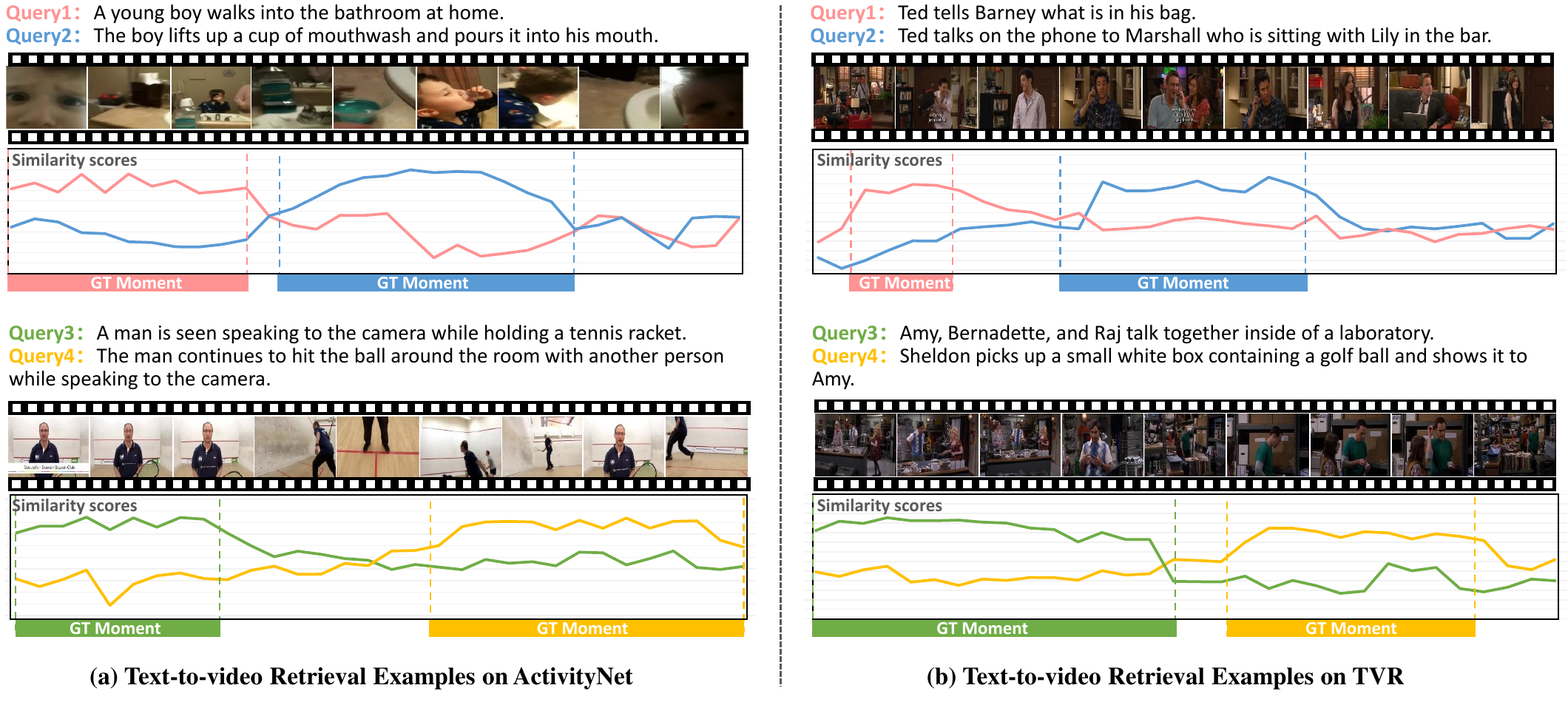}
    \vspace{-0.4cm}
    \caption{Visualization of text-to-video results on ActivityNet and TVR. In each block, we provide the query, Top-1 retrieved video, and text-clip similarity scores along the timeline. Dotted lines bound ground-truth (GT) moments for different queries. Note that GT moment intervals are for display only and are unavailable for training.}
    \vspace{-0.6cm}
    \label{fig:score}
\end{figure*}

\subsubsection{\textit{\textbf{Effect of Hyperparameters}}}
{In our model, $H$ is a key hyperparameter that determines the number of moment proposals the model generates, and also the attention head number in the masked multi-moment encoder. Generally, a larger $H$ allows the model to discover more moments within the video, increasing its capacity to capture all potential moments. However, larger $H$ also reduces the average duration of each moment, which can lead to incomplete representations of target moments.} To find a better trade-off, we study the effect of the number of $H=\{1, 2, 4, 8\}$. As shown in Table \ref{tab:numK}, the performance of our model reaches the peak with $H=4$ for videos in ActivityNet and TVR. This setting provides ample and distinguishable moment hints for retrieval.

In addition, we study the sensitivity of the loss coefficients $\lambda_{ret}$, $\lambda_{div}$, and $\lambda_{rel}$, on the ActivityNet and TVR datasets. Starting with the retrieval loss coefficient $\lambda_{ret}=0.02$, we vary $\lambda_{div}$ and $\lambda_{rel}$ over the values \{0.2, 0.4, 1, 2, 4\}. As shown in Fig. \ref{fig:para}, our model maintains robust performance across a range of hyperparameter values, with the optimal trade-off achieved at $\lambda_{div}=1$ and $\lambda_{rel}=1$ on both datasets. Each loss contributes to the retrieval performance, so keeping them within a similar order of magnitude ensures a balance between retrieval and moment learning objectives.

\subsubsection{\textit{\textbf{Alternative Span-to-mask Function}}} The focus of our work is the exploitation of moment-level modeling. During moment learning, it is flexible to adopt different span-to-mask transformations. Table \ref{tab:func} investigates three alternate window functions (\ie, Rectangular window, Triangular window, and Gaussian window \cite{wang2023gmmformer}). 
As can be seen, all three models achieve better than existing methods on both datasets, which demonstrates the effectiveness of active moment learning for PRVR. Besides, the Gaussian window slightly outperforms the Rectangular and Triangular windows. We attribute this to the smooth and natural characteristics of the Gaussian distribution \cite{wang2023gmmformer,zheng2022weakly}.
Unlike Rectangular and Triangular windows with sharp weight boundaries, the Gaussian window applies a gradually fading focus on frames farther from the center. This transition is beneficial in representing the natural progression of video moments, where frames near the center of an activity or event are often the most relevant to the query.

\begin{table}[t]
    \centering
    \caption{Performance with different span-to-mask functions on ActivityNet and TVR datasets. Our AMDNet shows consistent performance superiority.}
\label{tab:func}
    \renewcommand\arraystretch{1.1}	
    \resizebox{\columnwidth}{!}{ 
    \begin{tabular}{lcccccc} \toprule[1pt]
    \textbf{Dataset} &\textbf{Method} &\textbf{R@1} &\textbf{R@5} &\textbf{R@10} &\textbf{R@100} &\textbf{SumR} \\ \hline
    \multirow{3}{*}{ActivityNet} &Rectangular  &12.0 &32.3 &45.5 &81.9 &171.7 \\
    &Triangular &12.0 &\textbf{32.6} &45.7 &81.9 &172.2 \\  
    &Gaussian &\textbf{12.3} &{32.5} &\textbf{45.9} &\textbf{82.1} &\textbf{172.8} \\ \hline
    \multirow{3}{*}{TVR} &{Rectangular}  &{19.3} &{42.0} &{53.5} &{88.9} &{203.7} \\
     &{Triangular} &{19.1} &{42.3} &{54.0} &{88.9} &{204.3} \\  
    &{Gaussian} &{\textbf{19.7}} &{\textbf{42.4}} &{\textbf{54.1}} &{\textbf{88.9}} &{\textbf{205.1}}\\ 
    \bottomrule[1pt]
    \end{tabular}
    }
    \vspace{-0.3cm}
\end{table}

\subsubsection{\textit{\textbf{Effect of Model Scale}}}
To study the algorithm's scalability and performance across different model sizes, we experiment with the larger CLIP-ViT-L/14 backbone, increasing the overall model size from 152.17M to 428.63M parameters. As shown in Table \ref{tab:clip}, there are significant performance improvements with SumR increasing from 205.1 to 236.1 on the TVR dataset. The results demonstrate that our method scales effectively with larger models. In this work, we primarily validate our approach with CLIP-ViT-B/32, as it is widely used in video-text retrieval tasks \cite{dong2023dual,luo2022clip4clip,wu2023cap4video} and offers a trade-off between performance and computational efficiency.

\subsection{Qualitative Results}

\subsubsection{\textit{\textbf{Text-clip Similarity}}}
In this subsection, we investigate how the moment-enhanced video representation sensitively reacts to the text queries. As illustrated in Fig. \ref{fig:score}, we provide eight examples of text-to-video retrieval on both datasets, including the query, Top-1 retrieved video, and the fine-grained text-clip similarity scores. It can be found that: (1) given a specific query that only corresponds to a fragment of the video, our approach successfully retrieves the ground-truth video; (2) the similarity scores between the text-video pair exhibit clear moment boundaries, aligning well with the ground-truth moment. 
Take the first video as an example, our AMDNet returns the ground-truth video for Query1 and Query2. The similarity scores of the video with two queries distinguish different related moments. 
This suggests a sophisticated comprehension of the moment boundaries by our model.

\begin{table}[t]
    \centering
    \caption{Comparison of model size and retrieval performance using CLIP-ViT-B/32 and CLIP-ViT-L/14 on TVR dataset.}
\label{tab:clip}
    \renewcommand\arraystretch{1.1}	
    \resizebox{\columnwidth}{!}{ 
    \begin{tabular}{lcccccc} \toprule[1pt]
   \textbf{Backbone} &\textbf{\#Params} &\textbf{R@1} &\textbf{R@5} &\textbf{R@10} &\textbf{R@100} &\textbf{SumR} \\ \hline
    {CLIP-ViT-B/32} &{152.17M} &{{19.7}} &{{42.4}} &{{54.1}} &{{88.9}} &{{205.1}}\\ 
    {CLIP-ViT-L/14} &{428.63M } &{27.5} &{52.6} &{63.6} &{92.3} &{236.1} \\ 
    \bottomrule[1pt]
    \end{tabular}
    }
\end{table}

\begin{figure*}[t]
    \centering
    \includegraphics[width=0.9\textwidth]{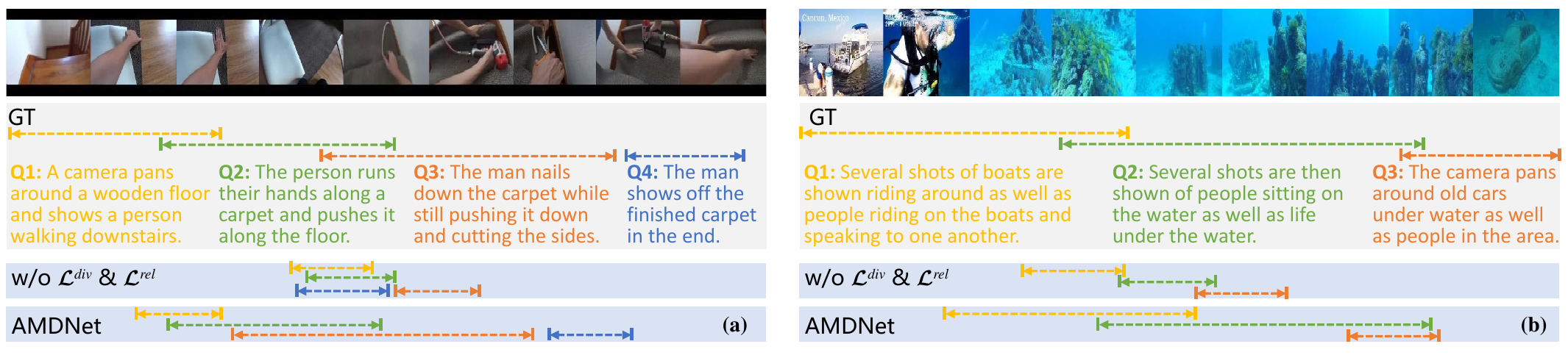}
    \vspace{-0.4cm}
    \caption{Qualitative comparison of the moment spans $({\bf c, w})$ predicted by AMDNet and the variant trained without $\mathcal{L}^{div}$ and $\mathcal{L}^{rel}$ on ActivityNet. We provide the GT moment spans for reference. The proposed moment optimization exhibits effectiveness in facilitating diversity and query-related moments.}
    \vspace{-0.3cm}
    \label{fig:cw}
\end{figure*}

\begin{figure}[t]
    \centering
    \includegraphics[width=0.8\columnwidth]{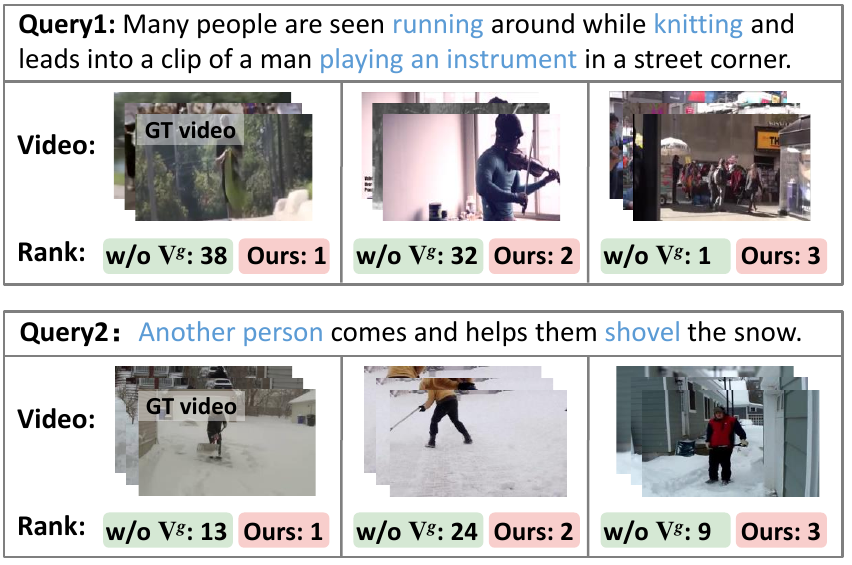}
    \vspace{-0.4cm}
    \caption{The text-to-video results on the ActivityNet test set. The ranking results are predicted by the baseline without activate moment discovering (denoted as ``w/o ${\bf V}^g$'') and our AMDNet, respectively. }
    \vspace{-0.3cm}
    \label{fig:tsne}
\end{figure}

\subsubsection{\textit{\textbf{Prediction of Moment Span}}}
Fig. \ref{fig:cw} shows some qualitative examples for moment prediction. In the gray rectangle, we indicate the GT moment spans for different queries using colorful dotted lines. In the blue rectangle, we provide the predicted moment spans by ``w/o $\mathcal{L}^{div}$ \& $\mathcal{L}^{rel}$'' and AMDNet, respectively. There are two observations in Fig. \ref{fig:cw}: (1) the prediction intervals of ``w/o $\mathcal{L}^{div}$ \& $\mathcal{L}^{rel}$'' are concentrated within similar ranges, particularly on videos containing multiple complex events. In contrast, AMDNet captures activities spanning different regions. (2) ``w/o $\mathcal{L}^{div}$ \& $\mathcal{L}^{rel}$'' fails to recognize the query-related moments, for instance, its predicted spans for Q1 and Q4 in Fig. \ref{fig:cw} (a) do not overlap with the GTs at all. AMDNet perceives the semantically related intervals to text query, proving useful moment hints. These visualizations further corroborate our superior results in Table \ref{tab:wo}.

\subsubsection{\textit{\textbf{Text-to-video Results}}}
We provide two examples of videos retrieved by our AMDNet and the baseline without activate moment discovering (``w/o ${\bf V}^g$'') in Fig. \ref{fig:tsne}. 
It can be found that introducing moment-based video grouping significantly improves the results of PRVR. For example, Query1 describes a complex moment involving multiple activities of \textit{running}, \textit{knitting}, and \textit{playing an instrument}. ``w/o ${\bf V}^g$'' is confused by videos containing similar activities, resulting in the GT video being ranked as low as 38th. By comparison, our approach successfully retrieves the GT video and ranks it 1st. 

\begin{figure}[t]
    \centering
    \includegraphics[width=0.8\columnwidth]{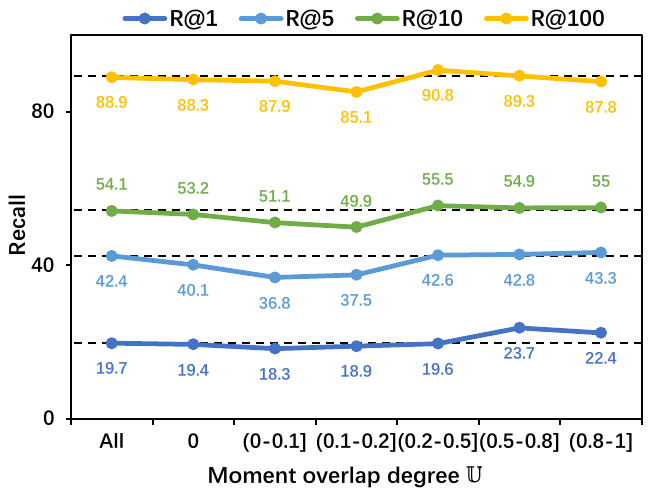}
    \vspace{-0.3cm}
    \caption{Text-to-video retrieval performance on queries with different degrees of moment overlap. Our model exhibits robust performance across different overlap settings.}
    \label{fig:overlap}
\end{figure}

Besides, we find that for challenging queries where relevant moments overlap, our model also performs well.
We define a \textit{moment overlap degree} $\mathbb{U}$$\in$[0,1] for each query as the maximum overlap between its relevant moment and other moments within the same video, and group test queries according to their $\mathbb{U}$ values. As shown in Fig. \ref{fig:overlap}, AMDNet exhibits robust performance across different overlap settings. Interestingly, the performance for queries with moderate to high overlap (\ie, $\mathbb{U}$$>$0.2) is competitive, or even better in some cases, compared to the overall performance on all queries.  
We think this is because overlapping moments provide additional semantic context that benefits PRVR.

\section{Limitations and Discussion}
{Although our approach sets the state-of-the-art in PRVR, there are still several limitations. As elaborated in the paper, we aim to highlight the key moments in untrimmed videos and estimate their accordance level with the given text query. Therefore, the proposed components expect given queries to maintain a meaningful context and describe distinguishable moments within the videos. If not, particularly for ambiguous queries corresponding to commonly occurring moments in the database, the retrieval ranking results may be affected. In the future, we are interested in exploring augmentations in the semantic context of queries and videos to improve robustness. 
}

\section{Conclusion}
This paper proposes a novel AMDNet for PRVR, which focuses on discovering and emphasizing semantically relevant video moments while suppressing redundant background content. Unlike existing methods that rely on multi-scale clip representations and suffer from content independence and information redundancy, our approach utilizes learnable span anchors and masked multi-moment attention to create more compact and informative video representations. We also introduce two loss functions--moment diversity loss and moment relevance loss--that enhance the model’s ability to distinguish between different moments and ensure alignment with text queries. These losses, in combination with a partially relevant retrieval loss, enable end-to-end optimization of our AMDNet. Our extensive experiments on large-scale datasets, including TVR and ActivityNet Captions, demonstrate the superior performance and efficiency of AMDNet.


\bibliographystyle{IEEEtran}
\bibliography{sample-base}

\begin{thebibliography}{10}
\providecommand{\url}[1]{#1}
\csname url@samestyle\endcsname
\providecommand{\newblock}{\relax}
\providecommand{\bibinfo}[2]{#2}
\providecommand{\BIBentrySTDinterwordspacing}{\spaceskip=0pt\relax}
\providecommand{\BIBentryALTinterwordstretchfactor}{4}
\providecommand{\BIBentryALTinterwordspacing}{\spaceskip=\fontdimen2\font plus
\BIBentryALTinterwordstretchfactor\fontdimen3\font minus \fontdimen4\font\relax}
\providecommand{\BIBforeignlanguage}[2]{{%
\expandafter\ifx\csname l@#1\endcsname\relax
\typeout{** WARNING: IEEEtran.bst: No hyphenation pattern has been}%
\typeout{** loaded for the language `#1'. Using the pattern for}%
\typeout{** the default language instead.}%
\else
\language=\csname l@#1\endcsname
\fi
#2}}
\providecommand{\BIBdecl}{\relax}
\BIBdecl

\bibitem{9878037}
X.~Wang, L.~Zhu, Z.~Zheng, M.~Xu, and Y.~Yang, ``Align and tell: Boosting text-video retrieval with local alignment and fine-grained supervision,'' \emph{IEEE Transactions on Multimedia}, vol.~25, pp. 6079--6089, 2023.

\bibitem{dong2022reading}
J.~Dong, Y.~Wang, X.~Chen, X.~Qu, X.~Li, Y.~He, and X.~Wang, ``Reading-strategy inspired visual representation learning for text-to-video retrieval,'' \emph{IEEE transactions on circuits and systems for video technology}, vol.~32, no.~8, pp. 5680--5694, 2022.

\bibitem{dong2021dual}
J.~Dong, X.~Li, C.~Xu, X.~Yang, G.~Yang, X.~Wang, and M.~Wang, ``Dual encoding for video retrieval by text,'' \emph{IEEE Transactions on Pattern Analysis and Machine Intelligence}, vol.~44, no.~8, pp. 4065--4080, 2021.

\bibitem{chen2020fine}
S.~Chen, Y.~Zhao, Q.~Jin, and Q.~Wu, ``Fine-grained video-text retrieval with hierarchical graph reasoning,'' in \emph{Proceedings of the IEEE/CVF Conference on Computer Vision and Pattern Recognition}, 2020, pp. 10\,638--10\,647.

\bibitem{miech2019howto100m}
A.~Miech, D.~Zhukov, J.-B. Alayrac, M.~Tapaswi, I.~Laptev, and J.~Sivic, ``Howto100m: Learning a text-video embedding by watching hundred million narrated video clips,'' in \emph{Proceedings of the IEEE/CVF International Conference on Computer Vision}, 2019, pp. 2630--2640.

\bibitem{liu2019use}
Y.~Liu, S.~Albanie, A.~Nagrani, and A.~Zisserman, ``Use what you have: Video retrieval using representations from collaborative experts,'' \emph{arXiv preprint arXiv:1907.13487}, 2019.

\bibitem{li2019w2vv++}
X.~Li, C.~Xu, G.~Yang, Z.~Chen, and J.~Dong, ``W2vv++ fully deep learning for ad-hoc video search,'' in \emph{Proceedings of the 27th ACM International Conference on Multimedia}, 2019, pp. 1786--1794.

\bibitem{faghri2017vse++}
F.~Faghri, D.~J. Fleet, J.~R. Kiros, and S.~Fidler, ``{VSE}++: Improving visual-semantic embeddings with hard negatives,'' in \emph{Proceedings of the British Machine Vision Conference}, 2018, pp. 935--943.

\bibitem{dong2019dual}
J.~Dong, X.~Li, C.~Xu, S.~Ji, Y.~He, G.~Yang, and X.~Wang, ``Dual encoding for zero-example video retrieval,'' in \emph{Proceedings of the IEEE/CVF Conference on Computer Vision and Pattern Recognition}, 2019, pp. 9346--9355.

\bibitem{dong2022partially}
J.~Dong, X.~Chen, M.~Zhang, X.~Yang, S.~Chen, X.~Li, and X.~Wang, ``Partially relevant video retrieval,'' in \emph{Proceedings of the 30th ACM International Conference on Multimedia}, 2022, pp. 246--257.

\bibitem{zhang2023multi}
G.~Zhang, J.~Ren, J.~Gu, and V.~Tresp, ``Multi-event video-text retrieval,'' in \emph{Proceedings of the IEEE/CVF International Conference on Computer Vision}, 2023, pp. 22\,113--22\,123.

\bibitem{nishimura2023large}
T.~Nishimura, S.~Nakada, and M.~Kondo, ``Large-scale vision-language models learn super images for efficient and high-performance partially relevant video retrieval,'' \emph{arXiv preprint arXiv:2312.00414}, 2023.

\bibitem{chen2023joint}
Z.~Chen, X.~Jiang, X.~Xu, Z.~Cao, Y.~Mo, and H.~T. Shen, ``Joint searching and grounding: Multi-granularity video content retrieval,'' in \emph{Proceedings of the 31st ACM International Conference on Multimedia}, 2023, pp. 975--983.

\bibitem{dong2023dual}
J.~Dong, M.~Zhang, Z.~Zhang, X.~Chen, D.~Liu, X.~Qu, X.~Wang, and B.~Liu, ``Dual learning with dynamic knowledge distillation for partially relevant video retrieval,'' in \emph{Proceedings of the IEEE/CVF International Conference on Computer Vision}, 2023, pp. 11\,302--11\,312.

\bibitem{wang2023gmmformer}
Y.~Wang, J.~Wang, B.~Chen, Z.~Zeng, and S.-T. Xia, ``Gmmformer: Gaussian-mixture-model based transformer for efficient partially relevant video retrieval,'' in \emph{Proceedings of the AAAI Conference on Artificial Intelligence}, 2024, pp. 5767--5775.

\bibitem{10054438}
S.~Huo, Y.~Zhou, R.~Wang, W.~Xiang, and S.-Y. Kung, ``Semantic relevance learning for video-query based video moment retrieval,'' \emph{IEEE Transactions on Multimedia}, vol.~25, pp. 9290--9301, 2023.

\bibitem{moon2023query}
W.~Moon, S.~Hyun, S.~Park, D.~Park, and J.-P. Heo, ``Query-dependent video representation for moment retrieval and highlight detection,'' in \emph{Proceedings of the IEEE/CVF Conference on Computer Vision and Pattern Recognition}, 2023, pp. 23\,023--23\,033.

\bibitem{wang2023ms}
J.~Wang, A.~Sun, H.~Zhang, and X.~Li, ``Ms-detr: Natural language video localization with sampling moment-moment interaction,'' \emph{arXiv preprint arXiv:2305.18969}, 2023.

\bibitem{momentdiff}
P.~Li, C.-W. Xie, H.~Xie, L.~Zhao, L.~Zhang, Y.~Zheng, D.~Zhao, and Y.~Zhang, ``Momentdiff: Generative video moment retrieval from random to real,'' in \emph{Advances in neural information processing systems}, 2023.

\bibitem{jiang2023progressive}
X.~Jiang, Z.~Chen, X.~Xu, F.~Shen, Z.~Cao, and X.~Cai, ``Progressive event alignment network for partial relevant video retrieval,'' in \emph{2023 IEEE International Conference on Multimedia and Expo (ICME)}, 2023, pp. 1973--1978.

\bibitem{ji2024multi}
Z.~Ji, Z.~Lin, H.~Wang, Y.~Pang, and X.~Li, ``Multi-task hierarchical convolutional network for visual-semantic cross-modal retrieval,'' \emph{Pattern Recognition}, vol. 151, p. 110398, 2024.

\bibitem{zhang2023consensus}
Y.~Zhang, Z.~Ji, Y.~Pang, and X.~Li, ``Consensus knowledge exploitation for partial query based image retrieval,'' \emph{IEEE Transactions on Circuits and Systems for Video Technology}, vol.~33, no.~12, pp. 7900--7913, 2023.

\bibitem{ji2024progressive}
Z.~Ji, J.~Wu, Y.~Wang, A.~Yang, and J.~Han, ``Progressive semantic reconstruction network for weakly supervised referring expression grounding,'' \emph{IEEE Transactions on Circuits and Systems for Video Technology}, 2024.

\bibitem{dong2018predicting}
J.~Dong, X.~Li, and C.~G. Snoek, ``Predicting visual features from text for image and video caption retrieval,'' \emph{IEEE Transactions on Multimedia}, vol.~20, no.~12, pp. 3377--3388, 2018.

\bibitem{li2024momentdiff}
P.~Li, C.-W. Xie, H.~Xie, L.~Zhao, L.~Zhang, Y.~Zheng, D.~Zhao, and Y.~Zhang, ``Momentdiff: Generative video moment retrieval from random to real,'' \emph{Advances in neural information processing systems}, vol.~36, 2024.

\bibitem{radford2021learning}
A.~Radford, J.~W. Kim, C.~Hallacy, A.~Ramesh, G.~Goh, S.~Agarwal, G.~Sastry, A.~Askell, P.~Mishkin, J.~Clark \emph{et~al.}, ``Learning transferable visual models from natural language supervision,'' in \emph{Proceedings of the 38th International Conference on Machine Learning}, 2021, pp. 8748--8763.

\bibitem{luo2022clip4clip}
H.~Luo, L.~Ji, M.~Zhong, Y.~Chen, W.~Lei, N.~Duan, and T.~Li, ``Clip4clip: An empirical study of clip for end to end video clip retrieval and captioning,'' \emph{Neurocomputing}, vol. 508, pp. 293--304, 2022.

\bibitem{wu2023cap4video}
W.~Wu, H.~Luo, B.~Fang, J.~Wang, and W.~Ouyang, ``Cap4video: What can auxiliary captions do for text-video retrieval?'' in \emph{Proceedings of the IEEE/CVF Conference on Computer Vision and Pattern Recognition}, 2023, pp. 10\,704--10\,713.

\bibitem{pei2023clipping}
R.~Pei, J.~Liu, W.~Li, B.~Shao, S.~Xu, P.~Dai, J.~Lu, and Y.~Yan, ``Clipping: Distilling clip-based models with a student base for video-language retrieval,'' in \emph{Proceedings of the IEEE/CVF Conference on Computer Vision and Pattern Recognition}, 2023, pp. 18\,983--18\,992.

\bibitem{liu2023revisiting}
R.~Liu, J.~Huang, G.~Li, J.~Feng, X.~Wu, and T.~H. Li, ``Revisiting temporal modeling for clip-based image-to-video knowledge transferring,'' in \emph{Proceedings of the IEEE/CVF Conference on Computer Vision and Pattern Recognition}, 2023, pp. 6555--6564.

\bibitem{deng2023prompt}
C.~Deng, Q.~Chen, P.~Qin, D.~Chen, and Q.~Wu, ``Prompt switch: Efficient clip adaptation for text-video retrieval,'' in \emph{Proceedings of the IEEE/CVF International Conference on Computer Vision}, 2023, pp. 15\,648--15\,658.

\bibitem{wang2024gmmformer}
Y.~Wang, J.~Wang, B.~Chen, T.~Dai, R.~Luo, and S.-T. Xia, ``Gmmformer v2: An uncertainty-aware framework for partially relevant video retrieval,'' \emph{arXiv preprint arXiv:2405.13824}, 2024.

\bibitem{hu2024maskable}
J.~Hu, D.~Guo, K.~Li, Z.~Si, X.~Yang, and M.~Wang, ``Maskable retentive network for video moment retrieval,'' in \emph{Proceedings of the 32nd ACM International Conference on Multimedia}, 2024, pp. 1476--1485.

\bibitem{yang2022video}
X.~Yang, S.~Wang, J.~Dong, J.~Dong, M.~Wang, and T.-S. Chua, ``Video moment retrieval with cross-modal neural architecture search,'' \emph{IEEE Transactions on Image Processing}, vol.~31, pp. 1204--1216, 2022.

\bibitem{lei2020tvr}
J.~Lei, L.~Yu, T.~L. Berg, and M.~Bansal, ``Tvr: A large-scale dataset for video-subtitle moment retrieval,'' in \emph{European Conference on Computer Vision}, 2020, pp. 447--463.

\bibitem{zhang2021video}
H.~Zhang, A.~Sun, W.~Jing, G.~Nan, L.~Zhen, J.~T. Zhou, and R.~S.~M. Goh, ``Video corpus moment retrieval with contrastive learning,'' in \emph{Proceedings of the 44th International ACM SIGIR Conference on Research and Development in Information Retrieval}, 2021, pp. 685--695.

\bibitem{hou2021conquer}
Z.~Hou, C.-W. Ngo, and W.~K. Chan, ``Conquer: Contextual query-aware ranking for video corpus moment retrieval,'' in \emph{Proceedings of the 29th ACM International Conference on Multimedia}, 2021, pp. 3900--3908.

\bibitem{10252035}
T.~Chen, W.~Wang, Z.~Jiang, R.~Li, and B.~Wang, ``Cross-modality knowledge calibration network for video corpus moment retrieval,'' \emph{IEEE Transactions on Multimedia}, vol.~26, pp. 3799--3813, 2024.

\bibitem{song2024emotional}
P.~Song, D.~Guo, X.~Yang, S.~Tang, and M.~Wang, ``Emotional video captioning with vision-based emotion interpretation network,'' \emph{IEEE Transactions on Image Processing}, vol.~33, pp. 1122--1135, 2024.

\bibitem{song2023emotion}
P.~Song, D.~Guo, X.~Yang, S.~Tang, E.~Yang, and M.~Wang, ``Emotion-prior awareness network for emotional video captioning,'' in \emph{Proceedings of the 31st ACM International Conference on Multimedia}, 2023, p. 589–600.

\bibitem{song2023contextual}
P.~Song, D.~Guo, J.~Cheng, and M.~Wang, ``Contextual attention network for emotional video captioning,'' \emph{IEEE Transactions on Multimedia}, vol.~25, pp. 1858--1867, 2023.

\bibitem{guo2024benchmarking}
D.~Guo, K.~Li, B.~Hu, Y.~Zhang, and M.~Wang, ``Benchmarking micro-action recognition: Dataset, methods, and applications,'' \emph{IEEE Transactions on Circuits and Systems for Video Technology}, vol.~34, no.~7, pp. 6238--6252, 2024.

\bibitem{HRNE}
P.~Pan, Z.~Xu, Y.~Yang, F.~Wu, and Y.~Zhuang, ``Hierarchical recurrent neural encoder for video representation with application to captioning,'' in \emph{Proceedings of the IEEE Conference on Computer Vision and Pattern Recognition}, 2016, pp. 1029--1038.

\bibitem{BAE}
L.~Baraldi, C.~Grana, and R.~Cucchiara, ``Hierarchical boundary-aware neural encoder for video captioning,'' in \emph{Proceedings of the IEEE Conference on Computer Vision and Pattern Recognition}, 2017, pp. 1657--1666.

\bibitem{PickNet}
Y.~Chen, S.~Wang, W.~Zhang, and Q.~Huang, ``Less is more: Picking informative frames for video captioning,'' in \emph{Proceedings of the European Conference on Computer Vision}, 2018, pp. 358--373.

\bibitem{OA-BTG}
J.~Zhang and Y.~Peng, ``Object-aware aggregation with bidirectional temporal graph for video captioning,'' in \emph{Proceedings of the IEEE Conference on Computer Vision and Pattern Recognition}, 2019, pp. 8327--8336.

\bibitem{ORG-TRL}
Z.~Zhang, Y.~Shi, C.~Yuan, B.~Li, P.~Wang, W.~Hu, and Z.-J. Zha, ``Object relational graph with teacher-recommended learning for video captioning,'' in \emph{Proceedings of the IEEE/CVF Conference on Computer Vision and Pattern Recognition}, 2020, pp. 13\,278--13\,288.

\bibitem{SAAT}
Q.~Zheng, C.~Wang, and D.~Tao, ``Syntax-aware action targeting for video captioning,'' in \emph{Proceedings of the IEEE/CVF Conference on Computer Vision and Pattern Recognition}, 2020, pp. 13\,096--13\,105.

\bibitem{chen2023weakly}
W.~Chen, G.~Li, X.~Zhang, S.~Wang, L.~Li, and Q.~Huang, ``Weakly supervised text-based actor-action video segmentation by clip-level multi-instance learning,'' \emph{ACM Transactions on Multimedia Computing, Communications and Applications}, vol.~19, no.~1, pp. 1--22, 2023.

\bibitem{chen2021cascade}
W.~Chen, G.~Li, X.~Zhang, H.~Yu, S.~Wang, and Q.~Huang, ``Cascade cross-modal attention network for video actor and action segmentation from a sentence,'' in \emph{Proceedings of the 29th ACM International Conference on Multimedia}, 2021, pp. 4053--4062.

\bibitem{ryu2021semantic}
H.~Ryu, S.~Kang, H.~Kang, and C.~D. Yoo, ``Semantic grouping network for video captioning,'' in \emph{proceedings of the AAAI Conference on Artificial Intelligence}, 2021, pp. 2514--2522.

\bibitem{gao2024audio}
J.~Gao, H.~Yang, M.~Gong, and X.~Li, ``Audio--visual representation learning for anomaly events detection in crowds,'' \emph{Neurocomputing}, vol. 582, p. 127489, 2024.

\bibitem{zhang2024descriptive}
B.~Zhang, J.~Gao, and Y.~Yuan, ``A descriptive basketball highlight dataset for automatic commentary generation,'' in \emph{Proceedings of the 32nd ACM International Conference on Multimedia}, 2024, pp. 10\,316--10\,325.

\bibitem{zheng2022weakly}
M.~Zheng, Y.~Huang, Q.~Chen, Y.~Peng, and Y.~Liu, ``Weakly supervised temporal sentence grounding with gaussian-based contrastive proposal learning,'' in \emph{Proceedings of the IEEE/CVF Conference on Computer Vision and Pattern Recognition}, 2022, pp. 15\,555--15\,564.

\bibitem{vaswani2017attention}
A.~Vaswani, N.~Shazeer, N.~Parmar, J.~Uszkoreit, L.~Jones, A.~N. Gomez, {\L}.~Kaiser, and I.~Polosukhin, ``Attention is all you need,'' \emph{Advances in neural information processing systems}, vol.~30, 2017.

\bibitem{chen2022multi}
W.~Chen, D.~Hong, Y.~Qi, Z.~Han, S.~Wang, L.~Qing, Q.~Huang, and G.~Li, ``Multi-attention network for compressed video referring object segmentation,'' in \emph{Proceedings of the 30th ACM International Conference on Multimedia}, 2022, pp. 4416--4425.

\bibitem{yang2024robust}
X.~Yang, J.~Zeng, D.~Guo, S.~Wang, J.~Dong, and M.~Wang, ``Robust video question answering via contrastive cross-modality representation learning,'' \emph{Science China Information Sciences}, vol.~67, no.~10, pp. 1--16, 2024.

\bibitem{yang2024learning}
X.~Yang, T.~Chang, T.~Zhang, S.~Wang, R.~Hong, and M.~Wang, ``Learning hierarchical visual transformation for domain generalizable visual matching and recognition,'' \emph{International Journal of Computer Vision}, pp. 1--27, 2024.

\bibitem{miech2020end}
A.~Miech, J.-B. Alayrac, L.~Smaira, I.~Laptev, J.~Sivic, and A.~Zisserman, ``End-to-end learning of visual representations from uncurated instructional videos,'' in \emph{Proceedings of the IEEE/CVF Conference on Computer Vision and Pattern Recognition}, 2020, pp. 9879--9889.

\bibitem{lin2017structured}
Z.~Lin, M.~Feng, C.~N.~d. Santos, M.~Yu, B.~Xiang, B.~Zhou, and Y.~Bengio, ``A structured self-attentive sentence embedding,'' \emph{arXiv preprint arXiv:1703.03130}, 2017.

\bibitem{krishna2017dense}
R.~Krishna, K.~Hata, F.~Ren, L.~Fei-Fei, and J.~Carlos~Niebles, ``Dense-captioning events in videos,'' in \emph{Proceedings of the IEEE International Conference on Computer Vision}, 2017, pp. 706--715.

\bibitem{yang2022vision}
J.~Yang, J.~Duan, S.~Tran, Y.~Xu, S.~Chanda, L.~Chen, B.~Zeng, T.~Chilimbi, and J.~Huang, ``Vision-language pre-training with triple contrastive learning,'' in \emph{Proceedings of the IEEE/CVF Conference on Computer Vision and Pattern Recognition}, 2022, pp. 15\,671--15\,680.

\bibitem{kingma2014adam}
D.~P. Kingma and J.~Ba, ``Adam: A method for stochastic optimization,'' \emph{arXiv preprint arXiv:1412.6980}, 2014.

\end{thebibliography}


\end{document}